\documentclass[sigconf]{acmart}
\renewcommand\footnotetextcopyrightpermission[1]{} %
\usepackage{multirow}
\usepackage[normalem]{ulem}
\useunder{\uline}{\ul}{}
\usepackage{bbm}
\usepackage{enumitem}
\usepackage{tablefootnote}
\usepackage{makecell}

\AtBeginDocument{%
  \providecommand\BibTeX{{%
    \normalfont B\kern-0.5em{\scshape i\kern-0.25em b}\kern-0.8em\TeX}}}

\begin{document}

\title{Road Planning for Slums via Deep Reinforcement Learning}

\author{Yu Zheng\footnotemark[1], Hongyuan Su\footnotemark[1], Jingtao Ding, Depeng Jin, Yong Li}
\thanks{*Both authors contributed equally to this research.}
\affiliation{%
  \institution{Department of Electronic Engineering, BNRist, Tsinghua University, Beijing, China}
  \city{}
  \country{}
  }
\email{liyong07@tsinghua.edu.cn}

\renewcommand{\shortauthors}{Zheng and Su, et al.}

\begin{abstract}

Millions of slum dwellers suffer from poor accessibility to urban services due to inadequate road infrastructure within slums, and road planning for slums is critical to the sustainable development of cities.
Existing re-blocking or heuristic methods are either time-consuming which cannot generalize to different slums, or yield sub-optimal road plans in terms of accessibility and construction costs.
In this paper, we present a deep reinforcement learning based approach to automatically layout roads for slums.
We propose a generic graph model to capture the topological structure of a slum, and devise a novel graph neural network to select locations for the planned roads.
Through masked policy optimization, our model can generate road plans that connect places in a slum at minimal construction costs.
Extensive experiments on real-world slums in different countries verify the effectiveness of our model, which can significantly improve accessibility by 14.3\% against existing baseline methods.
Further investigations on transferring across different tasks demonstrate that our model can master road planning skills in simple scenarios and adapt them to much more complicated ones, indicating the potential of applying our model in real-world slum upgrading.
The code and data are available at \textcolor{blue}{\url{https://github.com/tsinghua-fib-lab/road-planning-for-slums}}.

\end{abstract}


\maketitle

\section{Introduction}

With rapid urbanization, currently about 4 billion people around the world live in cities, while 1 billion of them live in over 200,000 slums~\cite{un2004challenge,wesolowski2010parameterizing}.
The vast majority of slums suffer from poor accessibility, with internal places not connected to external road systems, and many places not even having addresses~\cite{corburn2014informal,habitat2012streets}.
Besides being unreachable by motor vehicles, urban services depending on road systems, such as piped services of water and sanitation buried under roads, cannot be delivered to places in slums, which leads to severe problems in public health, urban environment, \textit{etc}~\cite{un2004challenge}.
To tackle these problems, local upgrading of slums has become the primary approach for the sustainable development of cities, rather than moving all the people to cities, due to the massive number of slum dwellers and the socio-economic costs~\cite{weru2004community,patel2012knowledge,habitat2012streets,andavarapu2013evolution}.
Particularly, improving the accessibility by planning roads plays an essential role in slum upgrading~\cite{un2014practical,brelsford2018toward}.

Different from city-level road planning which grows a road network from the top down and arranges land functionalities accordingly~\cite{farahani2013review}, road planning for slums is a bottom-up process in which existing houses determine the possible forms of the road network~\cite{un2014practical}.
Therefore, current city-level approaches cannot handle the \textit{micro-level} road planning within a slum.
Meanwhile, road planning for slums is challenging due to its large solution space.
Take a moderate-size slum as an example, the solution space of planning 40 road segments from 80 candidate locations surpasses $10^{23}$, which is too large for exhaustive enumeration.
In practical slum upgrading, re-blocking~\cite{mitlin2012urban,habitat2012streets} strategy is adopted.
It involves negotiations with multiple stakeholders and usually takes a long time for a specific case, thus it can not generalize globally to different slums.
Given the enormous number of slums, it is necessary to develop a computational method that can automatically accomplish road plans with superior connectivity at minimal construction costs~\cite{brelsford2018toward,un2014practical}.
Such a model can significantly benefit slum upgrading and eventually help achieve \textit{cities without slums}~\cite{boonyabancha2005baan,weru2004community,brelsford2018toward,habitat2012streets}.

One pioneering work by Brelsford \textit{et al.}~\cite{brelsford2018toward} formulates road planning for slums as a constrained optimization problem, and proposes a heuristic search method to generate road plans.
It makes this problem computationally solvable and has been adopted for slums in South Africa and India.
Although the heuristic can be applied to different slums, we empirically show that the quality of the obtained plans is not guaranteed, with the accessibility and construction costs far from optimal.
Fortunately, with the rapid development of artificial intelligence (AI), it is promising to leverage AI to solve the problem of road planning for slums.
First, data-driven parametric models have strong generalization ability, which can adapt to different scenarios~\cite{wang2022task,yang2022learning,kawaguchi2017generalization}.
In addition, AI models, especially deep reinforcement learning (DRL) algorithms, are good at searching in a large action space to optimize various objectives.
The action space can be effectively eliminated by predicting rewards with a value network and sampling actions via a policy network~\cite{konda1999actor,silver2016mastering,haarnoja2018soft,mnih2016asynchronous}.
Particularly, DRL has been deployed in similar planning tasks, such as solving the vehicle routing problem~\cite{nazari2018reinforcement,duan2020efficiently,zong2022rbg} and designing circuit chips~\cite{amini2022generalizable,roy2021prefixrl,mirhoseini2021graph}.

Inspired by the success of DRL, we propose a DRL-based method to solve this significant real-world problem, road planning for slums.
Since slums are diverse in the original geometric space, \textit{e.g.}, existing houses and paths can be in various irregular shapes, we propose a generic graph model to describe a slum, solving the problem from topology instead of geometry.
The topology invariance of the graph model makes our method capable of generalizing to different slums of arbitrary forms.
We further develop a policy network to select road locations and a value network to predict the performance of road planning based on a novel graph neural network (GNN), overcoming the difficulty of efficient search in the huge action space.
We design a topology-aware message passing mechanism for GNN, which first gathers various topological information to edges from nodes, faces, and edges themselves, then broadcasts edge embeddings back to learn effective representations of roads and places in the slum.
Furthermore, we develop a masked policy optimization method and connectivity-priority reward functions to optimize various objectives, including accessibility, travel distance, and construction costs.
We conduct experiments on real-world slums to verify the effectiveness of our proposed model.

To summarize, the contributions of this paper are as follows,

\begin{itemize}[leftmargin=*]
    \item We formulate road planning for slums as a sequential decision-making problem, and propose a DRL-based solution.
    \item We develop a novel GNN and a multi-objective optimization method based on a generic graph model for slums.
    The proposed model can learn effective representations of places and roads in a slum, which enables superior road planning policy.
    \item We conduct extensive experiments on slums in different countries, and the results demonstrate the advantage of our proposed method against baseline methods.
    Our model can generate road plans with both higher accessibility and lower construction costs.
    Moreover, we also show the transferability of our model from small slums to large slums, indicating the potential of applying our method in real-world slum upgrading.
\end{itemize}
\section{Problem Statement}\label{sec::prob}

\begin{figure}[t]
    \centering
    \includegraphics[width=0.99\linewidth]{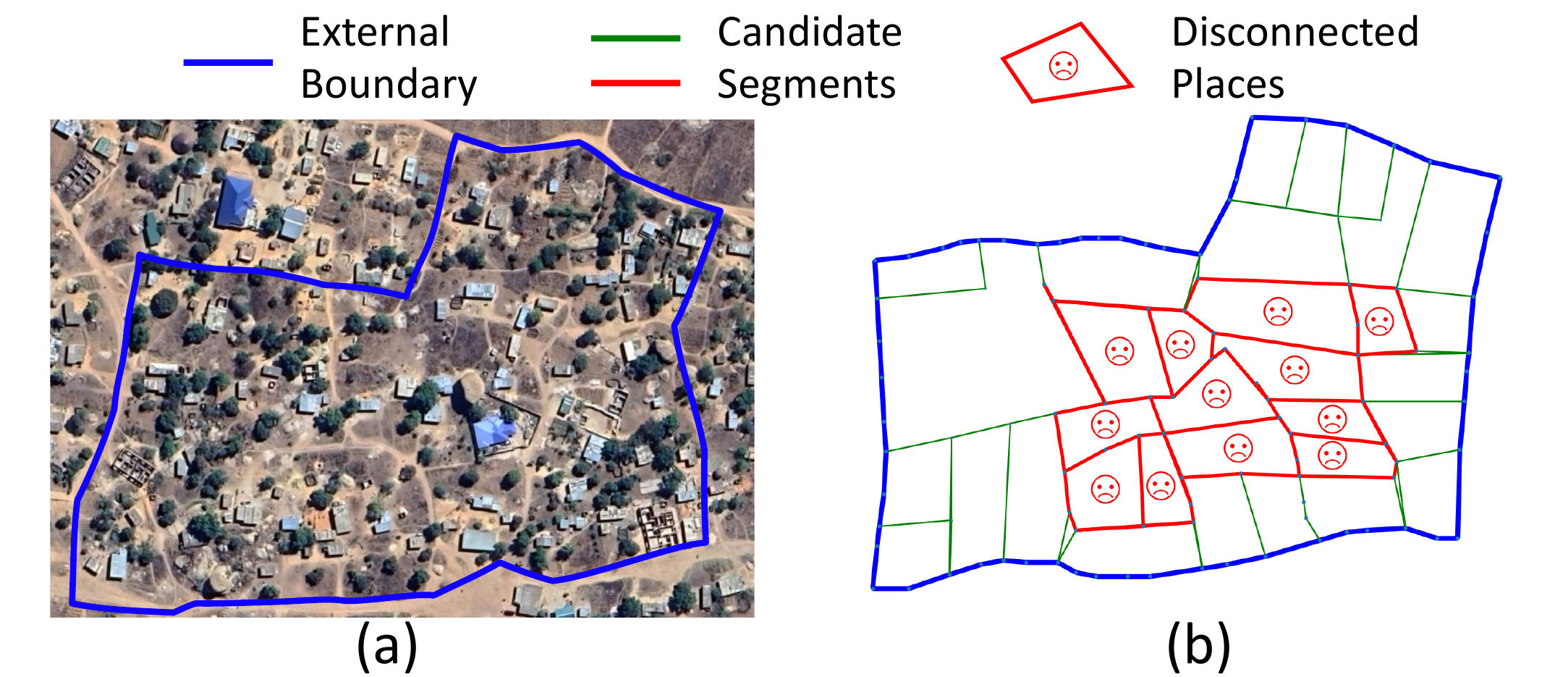}
    \vspace{-10px}
    \caption{
    (a) A slum in Harare, ZWE.
    Internal places in the slum are not connected to the external road system, making urban services inaccessible to slum dwellers.
    (b) Geometric description of the slum.
    Red polygons are places disconnected to roads, and internal segments (green and red) are candidate locations for new roads.
    Best viewed in color.
    }
    \vspace{-19px}
    \label{fig::prob_form}
\end{figure}

From the perspective of connectivity, a slum can be decomposed into two categories of elements, \textit{places} and \textit{roads}~\cite{brelsford2018toward}.
Specifically, places are the houses and internal facilities of the slum, and roads are the street system that connects various external urban services.
In most slums, a large fraction of places are disconnected from roads, as shown in Figure \ref{fig::prob_form}.
Such poor connectivity makes basic urban services inaccessible, \textit{e.g.}, ambulances and fire fighting trucks cannot reach the disconnected places during emergencies; water and sanitation pipes buried under roads cannot be provided.
Therefore, it is crucial to upgrade slums by planning more roads.
To deliver basic urban services, a minimal road network needs to make all places directly adjacent to roads, which is called universal connectivity~\cite{brelsford2018toward}.
Besides the minimally necessary accesses, more roads are expected to promote internal transportation and reduce travel distance for slum dwellers.
To minimize disruption to the slums, new roads are not allowed to pass through the middle of places, thus the candidate locations are restricted to the spacing between places. 
It is worth noting that each planned road segment also has a corresponding construction cost.

As illustrated in Figure \ref{fig::prob_form}(b), to describe the problem in geometric terms, a slum is a two-dimensional planar surface $U$ whose exterior boundaries $E$ are existing roads.
The surface is filled by a tessellation of faces (polygons) $P$, where each polygon $p_i$ is a place in the slum\footnote{A planar surface is a graph which can be drawn in the plane without any edges crossing. When a planar graph is drawn with no crossing edges, it divides the plane into a set of regions, called faces.}.
Polygon boundaries in the interior of the surface represent the spacing between places, which form a collection of segments $S$ and serve as the candidate locations for new roads.
Road planning is to select a subset of these segments for construction as roads.
Therefore, it can be formulated as follows:

\noindent\textbf{Input:} A planar surface $U$ with exterior boundaries $E$ for the slum, a collection of polygons $P$ for places in the slum, a collection of segments $S$ with their corresponding cost $C$ for road construction, and the road planning budget $K$.

\noindent\textbf{Output:} A subset $R$ of size $K$ from $S$ for construction as roads.

\noindent\textbf{Objective:} 
(1) Connecting all polygons in $P$ to the road system $E \cup R$. 
(2) Minimizing the travel distance between any pair of polygons $p_i$ and $p_j$ over the road network $E \cup R$ . 
(3) Minimizing the overall construction cost for the road plan $\sum_{i \in R}C_i$.

\section{Method}

\begin{figure}[t]
    \centering
    \includegraphics[width=0.99\linewidth]{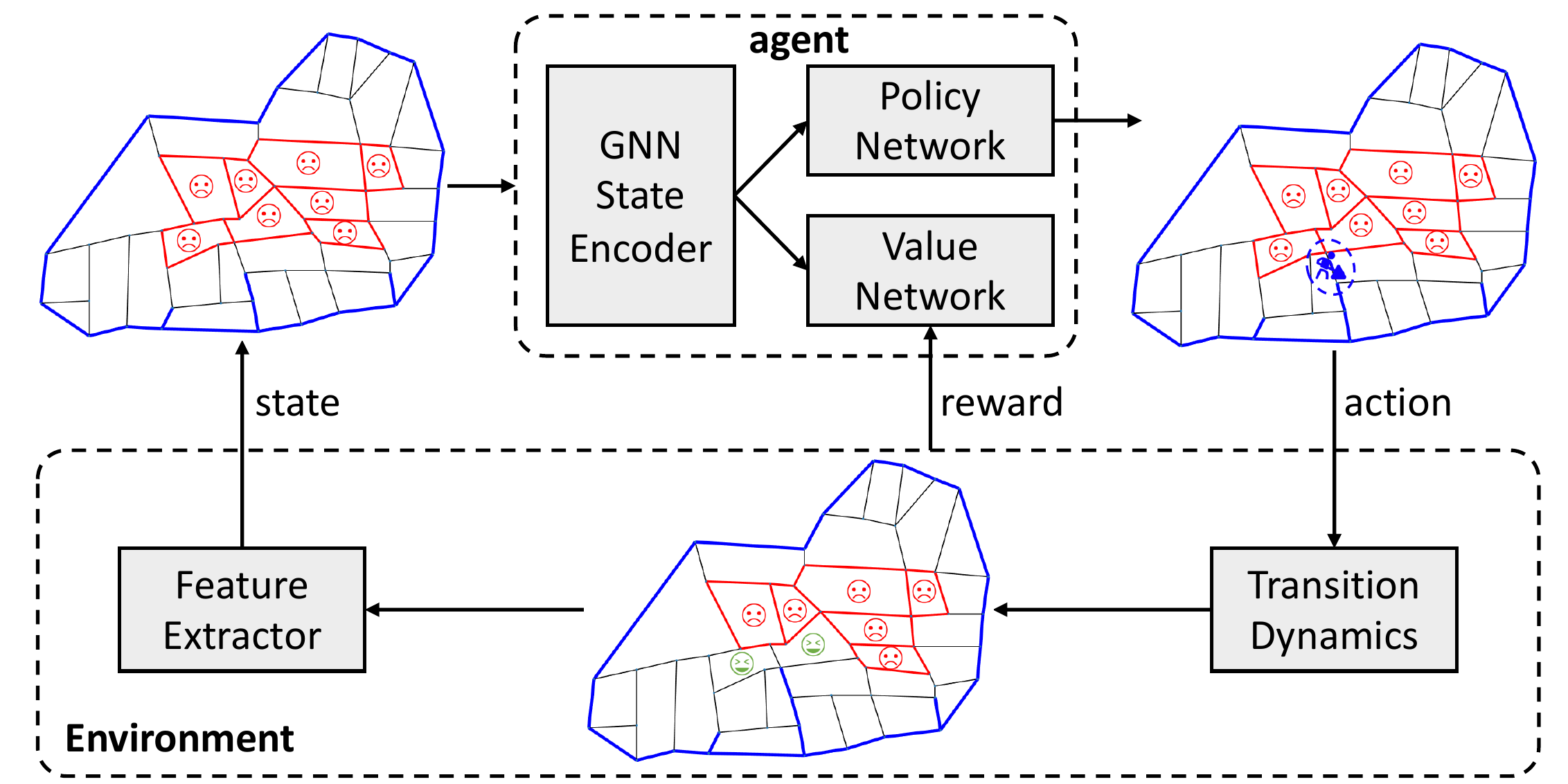}
    \vspace{-10px}
    \caption{
    Schematic of our approach.
    At each step, the agent receives states and rewards from the environment and outputs the road locations for the slum.
    Best viewed in color.
    }
    \vspace{-15px}
    \label{fig::rl}
\end{figure}

\subsection{Overall Framework}

\begin{figure*}[t]
    \centering
    \includegraphics[width=0.99\linewidth]{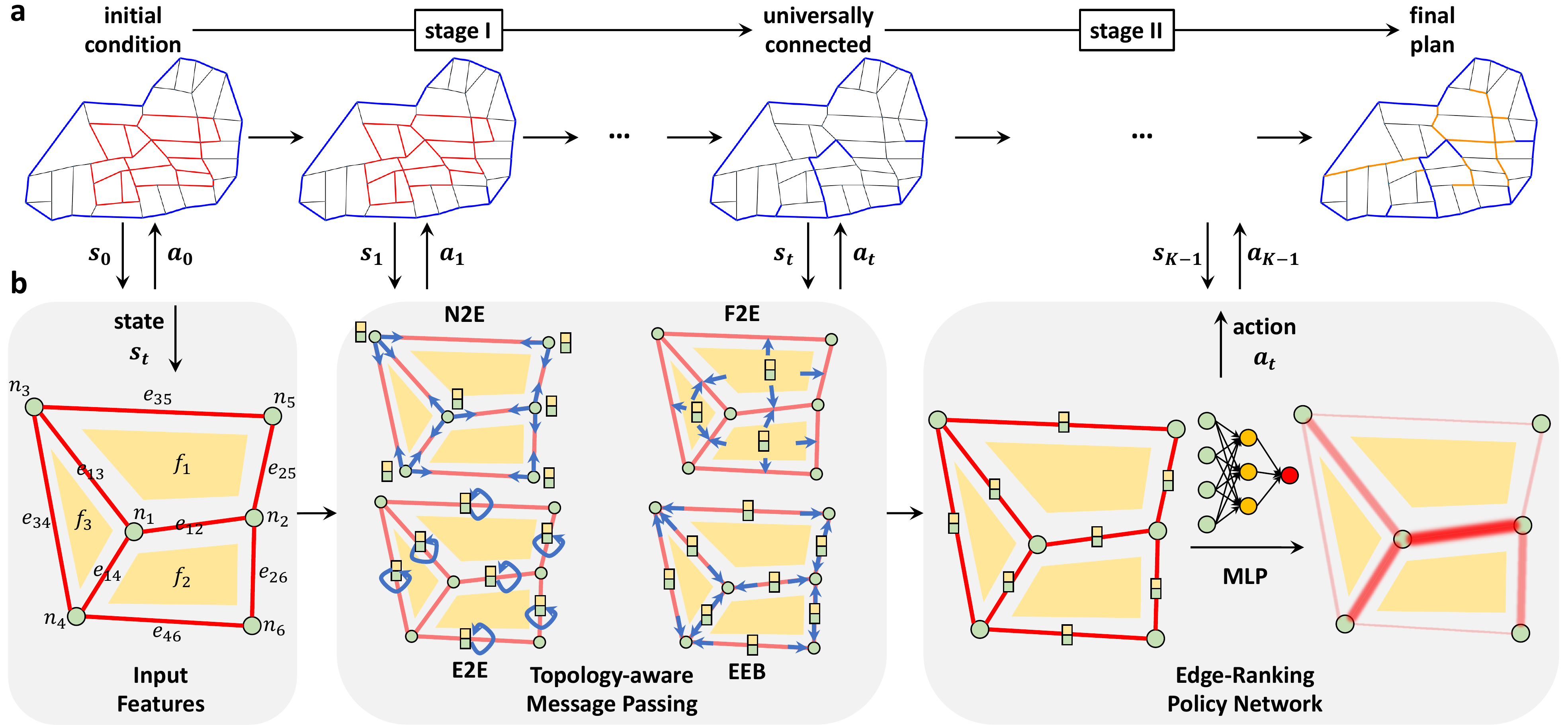}
    \vspace{-5px}
    \caption{
    (a) Road planning for slums as a sequential decision-making problem, where one single road segment is planed at each step.
    In stage \uppercase\expandafter{\romannumeral1}, the agent plan roads (blue) to achieve universal connectivity, \textit{i.e.}, all disconnected places are connected to the road system.
    In stage \uppercase\expandafter{\romannumeral2}, the agent add road segments (orange) to reduce travel distance.
    (b) The proposed GNN model.
    We design rich features for nodes, edges and faces.
    Topology-aware message passing is proposed, which contains Node2Edge Propagation (N2E), Face2Edge Propagation (F2E), Edge Self-Propagation (E2E) and Edge Embedding Broadcast (EEB).
    Finally, a edge-ranking policy network is developed to sample actions of edge selection.
    Best viewed in color.
    }
    \vspace{-5px}
    \label{fig::pipeline}
\end{figure*}

We formulate the road planning for slums as a sequential decision-making problem (see Section \ref{app::mdp} of the appendix for specific definitions of the Markov Decision Process (MDP)).
As illustrated in Figure \ref{fig::pipeline}a, given the planning budget, which is the total number of road segments, a road plan is accomplished through a sequence of location selection decisions, where at each step of the sequence, one new road segment is planned at a specific location.
The goal of the sequential decision-making problem is to improve the connectivity and accessibility of the slum at minimal costs.

As shown in Figure \ref{fig::rl}, we develop an agent with a policy network and a value network to take actions and predict returns, respectively, and a shared GNN model as the state encoder. 
To address the challenge of geometrical diversity, we tackle road planning for slums at the level of topology instead of geometry with a generic graph model (Section \ref{sec::graph_model}).
We then propose a novel GNN model to achieve a decent location selection policy on the graph (Section \ref{sec::gnn}).
In order to overcome the difficulty of multi-objective optimization in road planning, we further develop a masked policy optimization method with connectivity-priority reward functions (Section \ref{sec::optimization}).

\subsection{Graph Model}\label{sec::graph_model}

It is challenging to plan roads directly at the geometric level, since slums are very diverse in the original geometric space, \textit{e.g.}, the polygons of places can be in various irregular shapes, and the segments can intersect at almost any angle.
In addition, the spatial relationship between different geometries is more important for road planning than the specific shapes of geometries.
In contrast to the diverse geometries, there exists certain invariance in the topology of places and roads in cities~\cite{brelsford2018toward,xue2022quantifying}, which can support the uniform modeling of different slums.
Therefore, we solve the road planning problem from the topological viewpoint instead of the geometric one.
Specifically, we construct a planar graph to represent a slum with the contained places and roads, transforming the geometries into elements on the graph, such as nodes, edges, and faces.
In this way, we develop a generic graph model which can handle slums of arbitrary geometric forms at different scales with the same logic, solving the challenge of geometrical diversity.

\begin{figure}[t]
    \centering
    \includegraphics[width=0.99\linewidth]{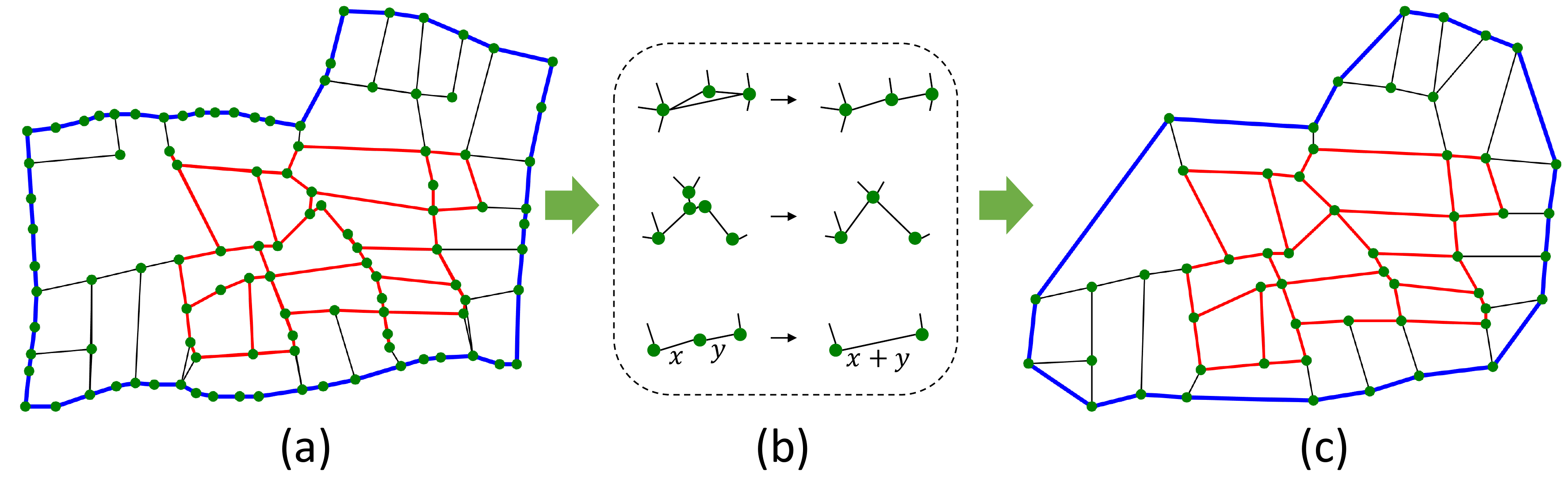}
    \vspace{-5px}
    \caption{
    (a) The constructed graph transformed from the original geometrical descriptions of the slum.
    Faces are polygons (places).
    Nodes are vertices of polygons.
    Edges are polygon boundary segments.
    (b) We simplify the graph by merging nearby edges (top) and nodes (middle) within certain threshold, and removing nodes with degree 2 (bottom).
    (c) The graph after preprocessing.
    Road planning solutions on the simplified graph can be easily mapped back to the original graph.
    Best viewed in color.
    }
    \vspace{-15px}
    \label{fig::preprocess}
\end{figure}

The planar graph is constructed based on the original geometrical descriptions of the slum, including the surface, polygons, and segments.
As shown in Figure \ref{fig::preprocess}(a), vertices and boundary segments of polygons become nodes and edges on the graph, respectively.
Meanwhile, the original polygons naturally become faces surrounded by edges on the planar graph, where each face in the graph represents a place which is usually a house in the slum.
Each edge has a \textit{road} attribute indicating whether it is a road segment or not, and a road segment can be either an existing external road or a planned new road.
Moreover, we preprocess the transformed planar graph of the slum to remove redundant information, as illustrated in Figure \ref{fig::preprocess}(b-c).
First, we merge multiple nodes/edges within a threshold distance as one node/edge, since they are supposed to share the same accessibility in the real space.
We then delete nodes with degree 2 and merge the corresponding two edges (construction costs are added) to simplify the graph, which have no influence on road planning.
Finally, we normalize the length of edges and align the coordinates, in order to support slums in different scales.

With the above generic graph model, road planning for slums is transformed into a sequential decision-making problem on a dynamic graph.
Specifically, states are the information of the current graph, and actions for a road planning policy are edge selections on the graph.
The graph also transits accordingly, \textit{i.e.}, the \textit{road} attribute of the selected edge changes from \textit{False} to \textit{True}, which in turn leads to subsequent changes in accessibility and travel distance of the slum.
For example, with the newly planned road, some faces (places) are connected to the road system, and the travel distance between several faces is reduced.
These changes are also reflected in the reward, which can be directly computed from the graph itself.

\subsection{Planning with Graph Neural Networks}\label{sec::gnn}

With the generic graph model of slums, we now introduce our proposed GNN model which performs road planning on the dynamic graph.
As the task is to select edges, a policy needs to decide the probability of choosing different edges at each step.
Since the topological information is critical to the effect of road planning, when computing the selection probability of each edge, it is necessary to consider its neighbors and even the whole graph, such as the travel distance of its neighboring faces.
Thus, we adopt GNN in our policy because of its strong ability to extract topological information and fuse neighborhood features.
As shown in Figure \ref{fig::rl}, we develop a GNN state encoder, which plays a fundamental role in the road planning agent.
The learned representations from GNN are shared between the policy network and the value network, serving as the basis for policy making and return prediction.

Slums exhibit complicated topological structure which can not be well captured by existing GNN models (see discussions in Section \ref{app::gnn_comparison} of the appendix).
To address the challenge of complex topology in road planning, we propose a novel GNN model which takes nodes, edges and faces into consideration.
Figure \ref{fig::pipeline}b demonstrates our proposed road planning policy based on GNN.
We first design rich features regarding accessibility, travel distance, and construction costs as the input of GNN.
We then design a topology-aware message passing mechanism to learn effective representations of topological elements on the graph.
Finally, we utilize an edge-ranking policy network to score edges based on the learned edge embeddings, supporting edge selection on the graph.

\begin{table}[t]
\caption{Designed features for topological elements.}
\vspace{-10px}
\label{tab::feature}
\begin{tabular}{cccc}
\toprule
\textbf{Topology}     & \textbf{Feature} & \textbf{Dimension} & \textbf{Type} \\
\midrule
\multirow{5}{*}{\bf{Node}} & Coordinates      & 2                  & Static        \\
                      & Centrality       & 4                  & Static        \\
                      & On Road          & 1                  & Dynamic       \\
                      & Road Ratio       & 1                  & Dynamic       \\
                      & Avg N2N Dis      & 1                  & Dynamic       \\ \hline
\multirow{3}{*}{\bf{Edge}} & Cost             & 1                  & Static        \\
                      & Road             & 1                  & Dynamic       \\
                      & Straightness     & 1                  & Dynamic       \\ \hline
\multirow{3}{*}{\bf{Face}} & Connected        & 1                  & Dynamic       \\
                      & Avg F2F Dis      & 1                  & Dynamic       \\
                      & F2E Dis          & 1                  & Dynamic       \\
\bottomrule
\end{tabular}
\vspace{-10px}
\end{table}

\subsubsection*{\textbf{Input Features for Topological Elements}}
Topological features reflect the current state of road planning, serving as the original input for GNN to learn representations of topological elements.
As illustrated in Table \ref{tab::feature}, we incorporate rich information about road planning into the designed features for nodes, edges, and faces.
Specifically, there are static features that do not change with the actions of the agent, such as the coordinates and construction cost, while most of the features are dynamic and alter according to actions at each step.
These meaningful features describe the current accessibility and travel distance of various places in the slum, which helps to decide which edges to plan as road segments.
For example, \textit{Connected} means whether a face is connected to the road system, thus building a road to an unconnected face can significantly improve the accessibility of the corresponding place to external urban services.
Similarly, \textit{Straightness} is the ratio of road network distance to the Euclidean distance of an edge, which directly indicates the travel distance between two places, and therefore selecting edges with high \textit{Straightness} can substantially reduce long detours in the slum.
These features support effective representation learning and subsequent decision-making, and details of all the designed topological features are introduced in Section \ref{app::feature} of the appendix.

\subsubsection*{\textbf{Topology-aware Message Passing}}
Since the policy selects edges on the graph to plan roads, we propose an edge-centric GNN to learn representations.
We first encode the input topological features to dense embeddings with separate weight matrices as follows,
\begin{equation}
    n_i^{(0)} = W_n^{(0)}A_{n_i},\quad
    e_{ij}^{(0)} = W_e^{(0)}A_{e_{ij}},\quad
    f_i^{(0)} = W_f^{(0)}A_{f_i},
\end{equation}
where $A_{n_i}$, $A_{e_{ij}}$ and $A_{f_i}$ are input attributes for nodes, edges and faces, $W_n$, $W_e$ and $W_f$ are learnable embedding matrices.

To address the challenge of complex topological elements, we design a topology-aware message passing mechanism, which first \textit{pulls} information from diverse topological elements into edges through \textit{node-to-edge propagation}, \textit{face-to-edge propagation}, and \textit{edge self-propagation}, and then \textit{pushes} aggregated topological information back through \textit{edge embedding broadcast}, as shown in Figure \ref{fig::pipeline}b.
The edge embeddings are obtained as follows.

\textit{Node2Edge Propagation.}
For each edge, we take the embeddings of its two connected nodes and propagate them through a linear transformation layer and a non-linear activation layer.
The node-to-edge message is computed as follows,
\begin{equation}
    e_{ij, n\rightarrow e}^{(l+1)} = \verb|tanh|(W_{n \rightarrow e}^{(l+1)}(n_i^{(l)} \| n_j^{(l)})),
\end{equation}
where $\|$ means concatenation, and $W_{n \rightarrow e}$ is a transformation layer.

\textit{Face2Edge Propagation.}
For each edge, we propagate the embeddings of its adjacent faces, and the face-to-edge message is computed as follows,
\begin{equation}
    e_{ij, f \rightarrow e}^{(l+1)} = \verb|tanh|(\frac{1}{N_{ij}^f}\sum_{k \in F_{ij}}{W_{f \rightarrow e}^{(l+1)}f_k^{(0)}}),\label{eq::face2edge}
\end{equation}
where $N_{ij}^f$ is the number of elements in $F_{ij}$, the set of adjacent faces for edge $e_{ij}$, and $W_{f \rightarrow e}$ is a linear transformation layer.

\textit{Edge Self-Propagation.}
Since each edge has its own attributes, we further include the propagation message from the edge itself, which is computed as follows,
\begin{equation}
    e_{ij, e \rightarrow e}^{(l+1)} = \verb|tanh|(W_{e \rightarrow e}^{(l+1)}e_{ij}^{(0)}),\label{eq::edge2edge}
\end{equation}
where a linear transformation matrix $W_{e \rightarrow e}$ is adopted.

The edge embedding is obtained by integrating the above three propagated messages as follows,
\begin{equation}
    e_{ij}^{(l+1)} = \verb|tanh|(W_e^{(l+1)}(e_{ij, n \rightarrow e}^{(l+1)} \| e_{ij, f \rightarrow e}^{(l+1)} \| e_{ij, e \rightarrow e}^{(l+1)})),
\end{equation}
where the three messages are concatenated and transformed with a linear layer $W_e^{(l+1)}$.

\textit{Edge Embedding Broadcast.}
We then \textit{push} the edge embeddings back to nodes to update their embeddings as follows,
\begin{align}
    &n_{i, e \rightarrow n}^{(l+1)} = \frac{1}{N_i}\sum_{j \in \mathcal{N}_i}e_{ij}^{(l+1)}, \\
    &n_{i}^{(l+1)} = n_{i}^{(l)} + n_{i, e \rightarrow n}^{(l+1)},
\end{align}
where for each node, we average the embeddings of its connected edges and add it to the node embedding.

By stacking multiple layers of the above topology-aware message passing, each node or edge can exchange information with neighbors on the graph.
We use the obtained embeddings at the last layer, $e_{ij}^{(L)}$ and $n_{i}^{(L)}$, as the final representations, where $L$ is a hyper-parameter in our model.
Through topology-aware message passing, the obtained edge representations can well capture the information about accessibility, travel distance, and construction costs of places and roads from its neighbors, which can effectively support the road planning policy.

\subsubsection*{\textbf{Edge-ranking Policy Network}}
The policy must generate the probability of selecting different edges at each step.
Therefore, we develop an edge-ranking policy network to score each edge, based on the obtained edge embeddings from GNN.
The score is calculated with a multi-layer perceptron (MLP) as follows,
\begin{equation}
    s_{ij} = \verb|MLP|_{p}(e_{ij}^{(L)}).\label{eq::logit}
\end{equation}
The action of edge selection is sampled from a probability distribution over different edges according to their corresponding scores $s_{ij}$ estimated by the policy network.
Since the obtained edge embeddings contain rich topological information, the road planning action made by the policy network takes into account the accessibility, travel distance, and construction cost of the slum.

\subsection{Multi-objective Policy Optimization}\label{sec::optimization}

Among the three objectives, accessibility, \textit{i.e.}, achieving universal connectivity for all places in the slum, is crucial for residents in the slum to access basic urban services, which is the primary target of road planning.
Therefore, it is necessary to prioritize connectivity when optimizing the policy, and further reduce travel distance after universal connectivity is achieved.
Meanwhile, for both connectivity and travel distance, it is desirable to optimize them at minimal construction cost.
Towards this end, we propose a masked policy optimization method and connectivity-priority reward functions with two stages, as shown in Figure \ref{fig::pipeline}a.
The optimization method encourages the policy to achieve universal connectivity in stage \uppercase\expandafter{\romannumeral1}, then reduce travel distance in stage \uppercase\expandafter{\romannumeral2}, preferring low construction cost in the whole process.

\subsubsection*{\textbf{Stage \uppercase\expandafter{\romannumeral1}}}
The goal of this stage is to achieve universal connectivity as quickly as possible, making all places in the slum connected to the road system and accessible to urban services.
Therefore, each new planned road segment is expected to connect more faces (places) that are not yet connected to any road segments.
Meanwhile, since road planning is a gradual extension of the existing road system, a new road segment can not be created as a separate component without touching the already planned roads.
We thus design an action mask to indicate feasible actions in this stage for the policy network, and the mask value of each edge is calculated as follows,
\begin{equation}
    m_{ij} = \mathbbm{1}[\verb|On_Road|(n_i)]\wedge \mathbbm{1}[\sum_{f \in F_{n_j}}{(1 - \verb|Connected|(f))} \geq 1],
\end{equation}
where $F_{n_j}$ represents all the faces that contain the node $n_j$.
In other words, the action mask requires the selected edge to start from a \textit{road node} and connect at least one \textit{unconnected face}
The mask value is multiplied over the obtained scores from the policy network in (\ref{eq::logit}), which serves as the selection probability of different edges,
\begin{equation}
    Prob(e_{ij}) = \frac{e^{s_{ij}}}{\sum_{(u,v) \in \mathcal{E}}{e^{s_{uv}}}}*m_{ij},\label{eq::prob}
\end{equation}
where $\mathcal{E}$ denotes all the edges on the graph.
With the action mask, only those edges that start from existing roads and connect disconnected faces will be considered by the policy.

Besides the action mask, we also design a corresponding reward function in this stage, which is a weighted sum of the number of newly connected faces and the construction cost of the planned road.
Given the action $a_k$ at the $k$-th step selecting the edge $e_{ij}$, the reward is calculated as follows,
\begin{equation}
    r_k = \alpha_1\sum_{f \in F_{n_j}}{(1 - \verb|Connected|(f))} + \alpha_2 \mathcal{C}_{e_{ij}},\label{eq::reward_1}
\end{equation}
where $\mathcal{C}_{e_{ij}}$ is the construction cost of the road segment specified by $e_{ij}$, and $\alpha_1$ and $\alpha_2$ are hyper-parameters in our model.

\begin{figure}[t]
    \centering
    \includegraphics[width=0.99\linewidth]{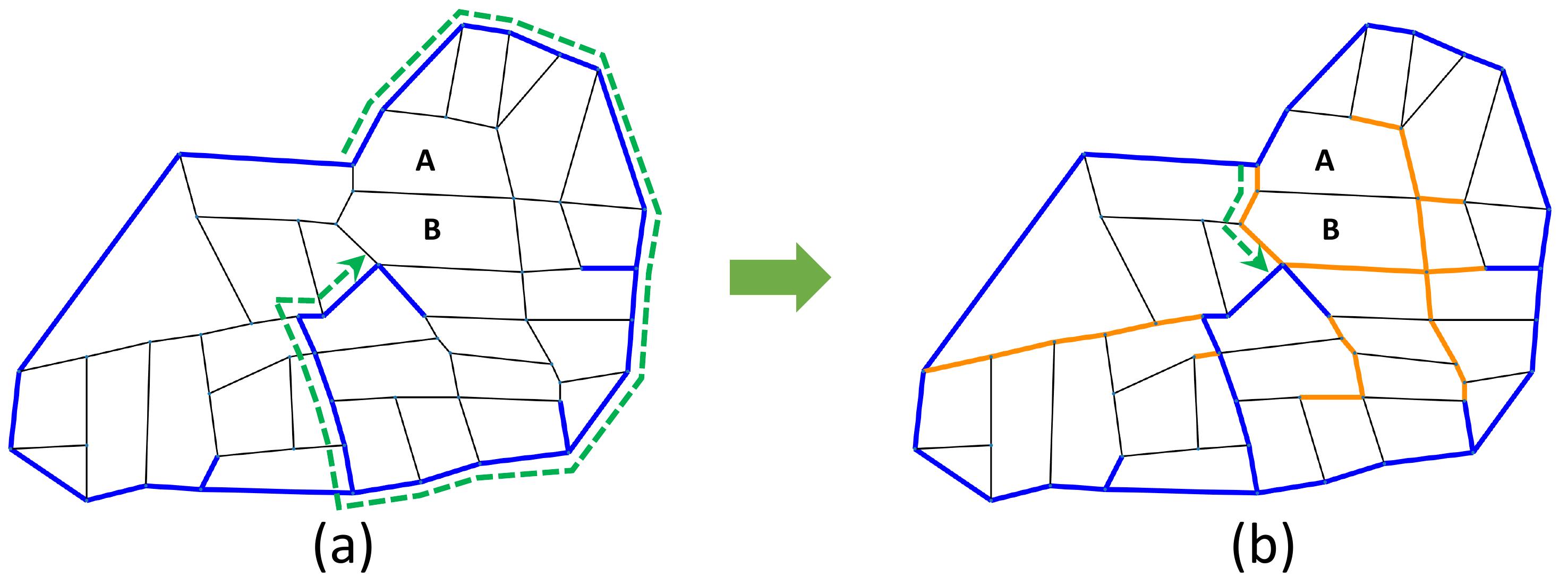}
    \vspace{-10px}
    \caption{
    (a) In stage \uppercase\expandafter{\romannumeral1}, the planned roads (blue segments) connect all disconnected places.
    However, many places, even nearby places, still suffer from high travel distance.
    In this example, place A and B are next to each other, while it requires a long detour (green path) between them by vehicle.
    (b) In stage \uppercase\expandafter{\romannumeral2}, roads are planned (orange segments) to reduce travel distance.
    Now the trip from place A to B (green path) is much shorter.
    Best viewed in color.
    }
    \vspace{-15px}
    \label{fig::detour}
\end{figure}

\subsubsection*{\textbf{Stage \uppercase\expandafter{\romannumeral2}}}
As shown in Figure \ref{fig::detour}(a), after the slum becomes universally connected in stage \uppercase\expandafter{\romannumeral1}, the generated road network looks like a tree with many dead-ends, which is undesirable in reality~\cite{alexander2019city,barthelemy2008modeling}.
Meanwhile, the traffic between some places is still poor and requires long detours, even for some nearby places.
Therefore, stage \uppercase\expandafter{\romannumeral2} aims to add more roads to reduce travel distance within the slum, as shown in Figure \ref{fig::detour}(b).
We still require the planned road segments to start from existing \textit{road nodes}, and the mask value of different edges are calculated as follows,
\begin{equation}
    m_{ij} = \mathbbm{1}[\verb|On_Road|(n_i)].
\end{equation}
The action probability is obtained in the same way as (\ref{eq::prob}).

For the reward function given action $a_k$ selecting edge $e_{ij}$, we compute the pairwise travel distance reduction of the slum, and combine it with construction cost,
\begin{align}
    &D(k) = \frac{|F|(|F| - 1)}{2}\sum_{u=1}^{|F|-1}\sum_{v=u+1}^{|F|}d(f_u, f_v; k),\\
    &r_k = \alpha_1(D(k) - D(k + 1)) + \alpha_2 \mathcal{C}_{e_{ij}},\label{eq::reward_2}
\end{align}
where $d(f_u, f_v; k)$ denotes the travel distance between two faces, $f_u$ and $f_v$, over the road network at the $k$-th step.

With the designed action mask and reward functions, the policy is guided to connect unconnected faces and reduce travel distance with low construction costs in the two stages, respectively.

\subsubsection*{\textbf{Value Network and Optimization.}}
Besides the policy network, we follow the actor-critic manner~\cite{konda1999actor} and develop a value network to predict the effect of road planning.
Since places and roads are captured with a graph, we compute graph-level representations to summarize the current state of the whole slum.
Specifically, we take the average of all the node embeddings and edges embeddings, and also include a one-hot encoding of the stage as follows,
\begin{align}
    &n_{avg} = \frac{1}{|\mathcal{N}|}\sum_{i=1}^{|\mathcal{N}|}{n_i^{(L)}}, e_{avg} = \frac{1}{|\mathcal{E}|}\sum_{(i, j) \in \mathcal{E}}{e_{ij}^{(L)}}, \label{eq::embed_avg}\\
    &h_g = n_{avg} \| e_{avg} \| \text{one-hot}(stage),
\end{align}
where $\mathcal{N}$ and $\mathcal{E}$ are the sets of nodes and edges, and $h_g$ is the graph representation.
We utilize an MLP model to predict the return,
\begin{equation}
    \hat{r} = \verb|MLP|_{v}(h_g).\label{eq::value}
\end{equation}
Finally, we adopt Proximal Policy Optimization (PPO)~\cite{schulman2017proximal} to update the parameters of the policy network and value network, which encourages the agent to conduct safe and efficient exploration in the action space.
Details of model training and inference are introduced in Section \ref{app::model} of the appendix.
\section{Experiments}

\subsection{Experiment Settings}

\subsubsection*{\textbf{Slum Data.}}
We conduct experiments on slums of different scales from different countries with publicly released data~\cite{brelsford2018toward}.
Table \ref{tab::data} shows the basic information of these slums, where we list the number of places and segments, as well as the size of the solution space.
Notably, all the slums suffer from poor accessibility, with over 40\% of places disconnected from road systems.
More details of the data are introduced in Section \ref{app::data} of the appendix.

\begin{table}[t]
\caption{Basic information of experimented slums. D.R. means the ratio of disconnected places.}
\vspace{-5px}
\label{tab::data}
\begin{tabular}{ccccc}
\toprule
\textbf{Location} & \textbf{Place} & \textbf{Segment} & \textbf{D.R.} & \textbf{Solution} \\
\midrule
Harare, ZWE & 32 & 85 & 37.5\% & $4\times 10^{20}$ \\
Cape Town, ZAF & 34 & 91 & 44.1\% & $6\times 10^{25}$ \\
Cape Town, ZAF & 59 & 164 & 59.3\% & $5\times 10^{47}$ \\
Mumbai, IND & 92 & 208 & 58.7\% & $1\times 10^{60}$ \\
\bottomrule
\end{tabular}
\vspace{-5px}
\end{table}

\begin{table*}[t]
\caption{Road planning performance comparison.
Lower is better.
F and INF means failing to achieve universal connectivity.
}
\vspace{-10px}
\label{tab::overall}
\begin{tabular}{c|ccc|ccc|ccc|ccc}
\toprule
\multirow{2}{*}{\textbf{Method}} & \multicolumn{3}{c|}{\textbf{Harare, ZWE}} & \multicolumn{3}{c|}{\textbf{Cape Town, ZAF (A)}} & \multicolumn{3}{c|}{\textbf{Cape Town, ZAF (B)}} & \multicolumn{3}{c}{\textbf{Mumbai, IND}} \\
                                 & \textbf{NR} & \textbf{AD} & \textbf{SC} & \textbf{NR}   & \textbf{AD}   & \textbf{SC}  & \textbf{NR}   & \textbf{AD}   & \textbf{SC} & \textbf{NR}   & \textbf{AD}   & \textbf{SC}  \\
\midrule
Random & 29 & 1.06 & 6.30 & F & INF & 10.83 & F & INF & 20.76 & F & INF & 26.05 \\
Random (masked) & 10 & 1.00 & 6.13 & 14 & 1.62 & 10.37 & 54 & 2.77 & 19.95 & 42 & 2.50 & 24.92 \\
\hline
Greedy-A (masked)$^*$ & 8$^*$ & 0.63 & 5.04 & 13$^*$ & 1.12 & 10.42 & 28$^*$ & 1.66 & 18.91 & 29$^*$ &  1.77 & 25.42 \\
Greedy-C$^*$ & 20 & 0.84 & 3.85$^*$ & 35 & 1.83 & 7.03$^*$ & F & INF & 14.10$^*$ & F & INF & 19.45$^*$ \\
Greedy-C (masked)$^*$ & 11 & 0.84 & 3.85$^*$ & 14 & 1.81 & 7.23$^*$ & 35 & 2.22 & 14.29$^*$ & 45 &  2.81 & 19.28$^*$ \\
\hline
GAN (manually adjusted) & - & 0.70 & 5.71 & - & 1.33 & 9.52 & - & 2.05 & 17.72 & - & 1.72 & 24.34 \\
VAE (manually adjusted) & - & 0.71 & 5.14 & - & 1.31 & 10.70 & - & 2.06 & 17.58 & - & 1.68 & 23.84 \\
\hline
MST (masked) & \uline{11} & 0.59 & 5.57 & \uline{14} & 1.17 & 8.75 & 35 & \uline{1.54} & 17.16 & 45 & 1.63 & \uline{22.92} \\
GA-G (masked) & \uline{11} & 0.58 & \uline{4.60} & \uline{14} & 1.14 & 8.72 & 34 & 1.99 & 18.95 & 42 & 1.87 & 24.26 \\
GA-S (masked) & - & 0.58 & 5.25 & - & 1.21 & 8.44 & - & 1.89 & 17.72 & - & 1.88 & 23.22 \\
HS-MC (masked) & 13 & 0.62 & 5.31 & 16 & 1.09 & 9.09 & 37 & 1.55 & 16.98 & 43 & 1.61 & 23.00 \\
\hline
DRL-MLP (ours, masked) & \uline{11} & \uline{0.52} & \bf{4.38} & \uline{14} & \uline{0.96} & \uline{8.28} & \uline{32} & 1.57 & \uline{15.66} & \uline{31} & \uline{1.52} & 22.93    \\
DRL-GNN (ours, masked) & \bf{9} & \bf{0.50} & \uline{4.60} & \bf{13} & \bf{0.93} & \bf{8.24} & \bf{31} & \bf{1.51} & \bf{15.62} & \bf{29} & \bf{1.51} & \bf{22.82} \\
impr\% v.s. HS-MC & -25.0\% & -19.4\% & -17.5\% & -18.8\% & -14.7\% & 9.8\% & -16.2\% & -2.6\% & -8.0\% & -32.6\% & -6.21\% & -0.8\% \\
\hline
Build All Roads & - & 0.47 & 11.50 & - & 0.80 & 19.82 & - & 1.21 & 37.55 & - & 1.36 & 49.25 \\
\bottomrule
\multicolumn{13}{l}{\makecell[l]{\small $^*$ Although they are equal to or even smaller than the \textbf{bolded} numbers, these methods exhibit imbalanced results with much worse performance on\\\small the other two metrics. Thus, the \textbf{bolded} and \uline{underlined} numbers are assigned to \textbf{the lowest} and \uline{the second lowest} values, \textit{excluding greedy methods}.}} \\
\end{tabular}
\vspace{-5px}
\end{table*}

\subsubsection*{\textbf{Baselines.}}
We compare our model with the following methods.
\begin{itemize}[leftmargin=*]
    \item \textbf{Random.}
    This method selects road segments randomly.
    \item \textbf{Greedy.}
    This method selects new road segments greedily according to accessibility (Greedy-A) and construction cost (Greedy-C).
    \item \textbf{Masked.}
    We add our proposed action mask to Random and Greedy baselines.
    Masked baselines select road segments that are True in the mask randomly (greedily).
    \item \textbf{Minimum Spanning Tree (MST).}
    A graph is built where nodes represent slums, edges represent road segments and edge weights represent road construction costs.
    We use Kruskal's algorithm~\cite{kruskal1956shortest} to grow a minimum spanning tree.
    \item \textbf{Genetic Algorithm (GA)~\cite{gad2021pygad}.}
    This type of method is widely adopted in road planning.
    We include a generative version (GA-G) that adopts a linear layer as genes and builds one road at one step by multiplying edge features with a linear layer as sampling probability.
    We also include a swap version (GA-S) that directly uses the selection of road segments as genes and performs swapping between different solutions at each step.
    \item \textbf{Heuristic Search (HS-MC)~\cite{brelsford2018toward}.}
    This recently proposed method formulates road planning for slums as a constrained optimization problem.
    It samples paths from external boundary roads to unconnected places using the Monte Carlo techniques~\cite{binder1993monte}.
    \item \textbf{DRL-MLP.}
    We implement a simplified DRL model by replacing the proposed GNN with an MLP, thus it ignores topological information when planning roads. 
\end{itemize}
It is worthwhile to notice that Greedy-A, MST, GA, HS-MC and our DRL models are all with action masks themselves.
We also include two generative models~\cite{fang2022incorporating,kempinska2019modelling}, based on Generative Adversarial Networks (GAN)~\cite{goodfellow2020generative} and Variational Auto-Encoder (VAE)~\cite{kingma2013auto}, though manual adjustments are required for these methods.
Details of all the baselines are introduced in Section \ref{app::baseline} of the appendix.

\subsubsection*{\textbf{Evaluation Metrics.}}
As introduced in Section \ref{sec::prob}, we evaluate a road plan concerning accessibility, travel distance, and construction cost.
The specific definitions are as follows,
\begin{itemize}[leftmargin=*]
    \item For accessibility, it is desired to achieve universal connectivity as early as possible, thus we calculate the \uline{n}umber of \uline{r}oad segments (NR) consumed to achieve universal connectivity.
    \item For travel distance, we compute the \uline{a}verage \uline{d}istance (AD) between any pair of places in the slum over the road network.
    \item We define the construction cost of each road segment as its length, and calculate the \uline{s}um of \uline{c}osts (SC) of all planned roads.
\end{itemize}
It is worth noting that all the metrics are \textit{the lower the better}.

\subsubsection*{\textbf{Model Implementation.}}
We implement the proposed model with PyTorch~\cite{paszke2019pytorch}, and all the codes and data to reproduce the results in this paper are released at \textcolor{blue}{\url{https://github.com/tsinghua-fib-lab/road-planning-for-slums}}.
We implement the greedy and GA baselines and integrate them into our framework.
For the heuristic search baseline, we use the codes released in~\cite{brelsford2018toward}.
We carefully tune the hyper-parameters of our model, including learning rate, regu larization, \textit{etc}.
For each road planning task, we collect millions of samples and train our model on a single server with an Nvidia GeForce 2080Ti GPU, which usually takes about 2 hours.
A full list of hyper-parameters is provided in Section \ref{app::our_model} of the appendix.

\subsection{Performance Comparison}\label{sec::perf_comp}

We set the planning budget (episode length) as 50\% of the number of candidate segments.
Results of our model and baselines are illustrated in Table \ref{tab::overall}, where we also include a reference model (\textit{Build All Roads}) that sets 100\% of candidate segments as roads.
NR is not applicable to GA-S since it is not a generative method.
From the results, we have the following observations,
\begin{itemize}[leftmargin=*]
    \item \textbf{Random and greedy algorithms are ineffective for road planning.}
    Randomly choosing locations fails to achieve universal connectivity in all slums except for the smallest one.
    Greedy-C achieves the lowest construction cost for all four slums, while it fails to achieve universal connectivity in the two largest slums.
    Adding action masks can help these methods to achieve universal connectivity, however, the travel distance is still the worst.
    Similarly, Greedy-A is the earliest to achieve universal connectivity, however, the construction cost is the worst, and the travel distance is also much worse than other methods.
    Thus we do not consider these trivial methods in the following comparisons.
    \item \textbf{Generative models are not suitable to road planning for slums.}
    As stated in Section \ref{app::baseline} of the appendix, to obtain road plans for slums with the two generative models~\cite{fang2022incorporating,kempinska2019modelling}, much of the work has to be conducted manually by human labor, which betrayed our original intention to automate the process of road planning.
    Not surprisingly, since they are not suitable to the sequential decision-making task, the performance of GAN and VAE falls far behind our proposed method and the HS-MC baseline.
    More discussions can be found at Section \ref{app::difference_generative} of the appendix.
    \item \textbf{DRL-based methods have significant advantages over other approaches.}
    DRL-MLP and DRL-GNN outperform GA-G, GA-S, and HS-MC on all metrics.
    The two DRL-based methods achieve much better road planning performance, with average reductions of about 23.2\%, 10.7\%, and 9.0\% in NR, AD, and SC over the four slums.
    Baselines like GA and HS-MC fail to explore the solution space efficiently, making it difficult to obtain high-quality road plans.
    The performance gap verifies the strong ability of DRL to optimize multiple objectives in a large action space.
    \item \textbf{Our proposed model achieves the best performance.}
    Regarding accessibility, our model is the fastest to achieve universal connectivity for all slums, which is critical under tight planning budgets.
    Compared with HS-MC, our model connects all places with 3 fewer road segments (NR) for slums in Harare and Cape Town, and 14 fewer road segments in the largest slum in Mumbai, IND.
    Meanwhile, with respect to travel distance and construction cost, our model consistently outperforms baseline methods.
    Specifically, our method reduces AD by 19.4\% and 14.7\% for slums in Harare and Cape Town, respectively, and reduces SC by 11.1\% for slums in Cape Town.
    In particular, the road plan obtained by our method achieves a travel distance very close to that of \textit{Build All Roads} at a much lower cost, making it more economical in real slum upgrading.
    Our model can capture topological information through the generic graph model and the novel GNN, and perform efficient searches in the large action space via masked policy optimization.
    These special designs enable our model to achieve superior performance in road planning for slums.
\end{itemize}

\begin{figure}[t]
    \centering
    \includegraphics[width=0.99\linewidth]{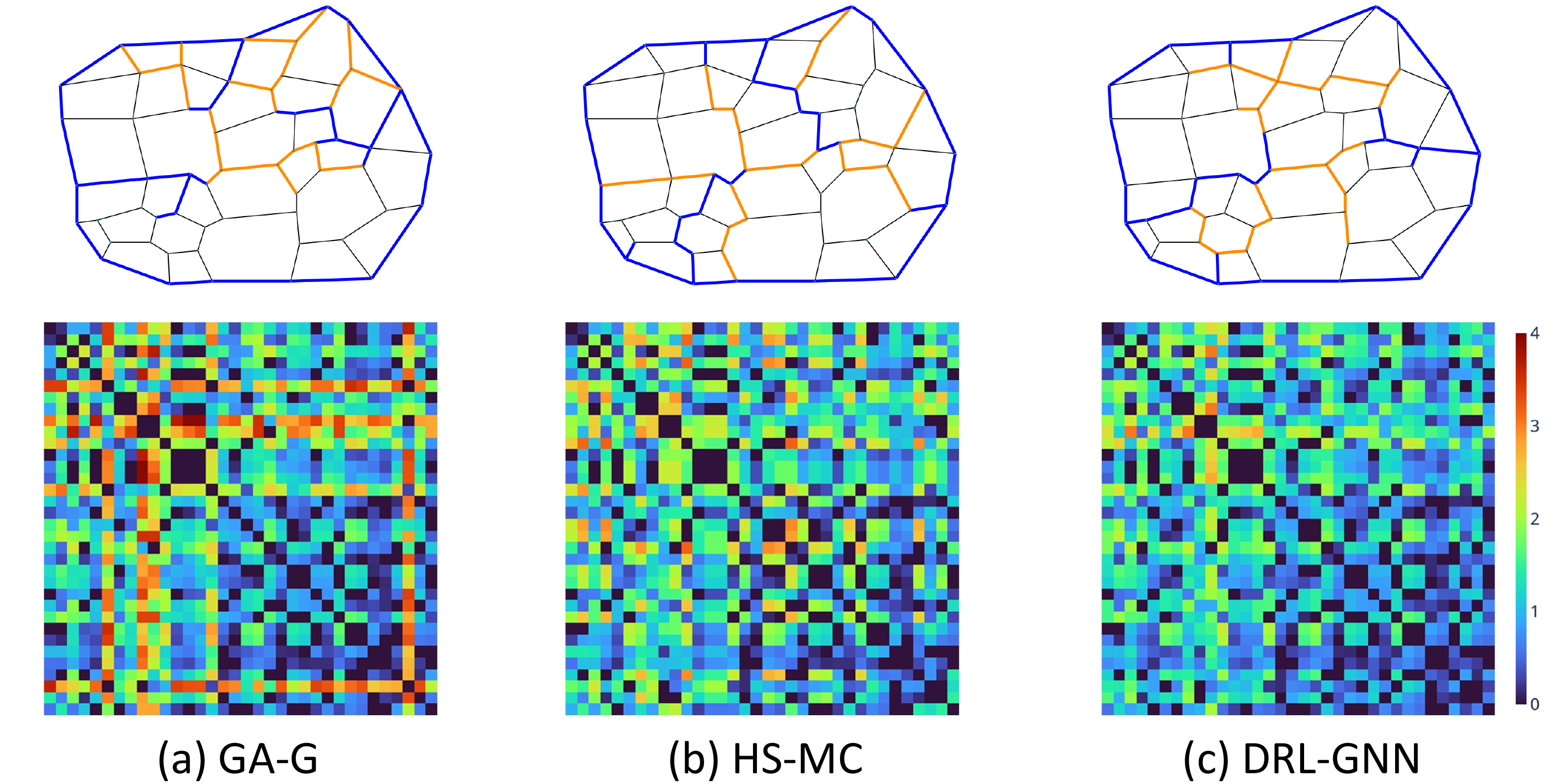}
    \vspace{-10px}
    \caption{
    The generated road plans for the slum in Cape Town, ZAF, and their corresponding travel distance matrices of (a) GA (b) HS-MC (c) DRL-GNN.
    Best viewed in color.
    }
    \vspace{-10px}
    \label{fig::plan_distance}
\end{figure}

Figure \ref{fig::plan_distance} demonstrates the generated road plans of different models for the slum in Cape Town, ZAF, and their corresponding travel distance matrices.
Although universal connectivity is achieved in all plans, the travel distance varies significantly across different methods.
In the road plans of baselines, slum dwellers in some places have to travel a long detour to reach each other, which corresponds to several \textit{hot} regions in the travel distance matrices as shown in Figure \ref{fig::plan_distance}(a-b).
In contrast, our method utilizes the progress in travel distance as the reward and optimizes it in stage \uppercase\expandafter{\romannumeral2}.
Specifically, the proposed GNN model can detect places that suffer from long detours through topology-aware message passing on the graph, and add targeted roads to reduce travel distance effectively.
Thus there are much fewer \textit{hot} regions as shown in Figure \ref{fig::plan_distance}(c).
In addition, as demonstrated in Figure \ref{fig::plan_distance}, our model is able to grow a road network in a less costly way, with the total length of planned roads much shorter than baselines by about 10\%.
We provide the complete road plans of all methods for all slums in Section \ref{app::plan} of the appendix.

Our proposed model can reach convergence in less than 100 iterations, which usually takes only about 2 hours.
We provide a visual plot of the convergence of our model in Section \ref{app::convergence} of the appendix.
One alternative way to accomplish road planning for slums is to set the total construction cost as the planning budget, instead of the number of road segments, and we provide the corresponding results in Section \ref{app::cost_budget} of the appendix.

\begin{figure}[t]
    \centering
    \includegraphics[width=0.99\linewidth]{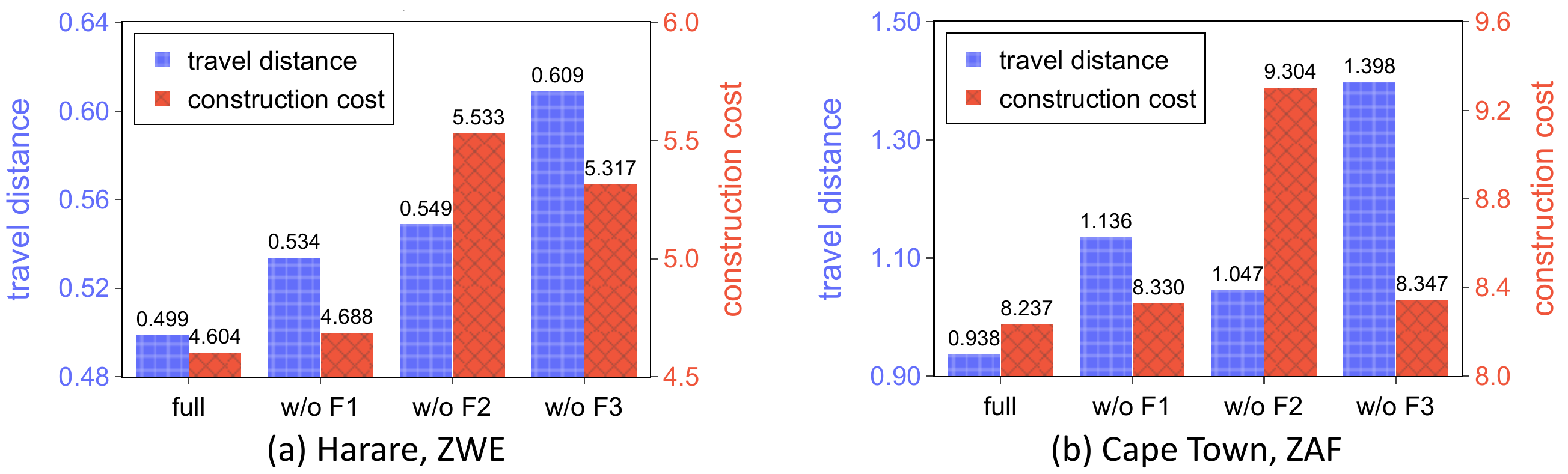}
    \vspace{-10px}
    \caption{
    Performance of DRL-GNN and its variants that remove different features, including Centrality (F1), Road (F2) and Straightness (F3) for slums in (a) Harare, ZWE (b) Cape Town, ZAF.
    Best viewed in color.
    }
    \vspace{-10px}
    \label{fig::input_features}
\end{figure}

\begin{figure}[t]
    \centering
    \includegraphics[width=0.99\linewidth]{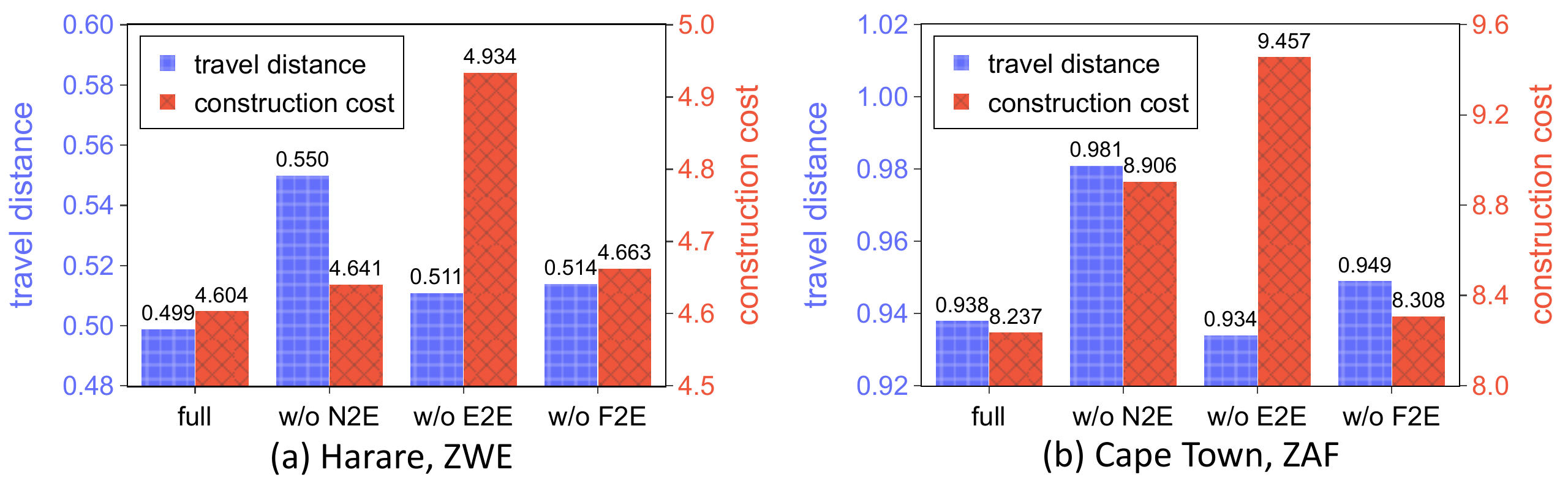}
    \vspace{-10px}
    \caption{
    Performance of DRL-GNN and its variants that remove node2edge propagation (N2E), edge self-propagation (E2E) and face2edge propagation (F2E) for slums in (a) Harare, ZWE (b) Cape Town, ZAF.
    Best viewed in color.
    }
    \vspace{-10px}
    \label{fig::message_passing}
\end{figure}

\subsection{Ablation Study}\label{sec::ablation}

\subsubsection*{\textbf{Graph Modeling}}
The spatial topological relationships between places and roads in a slum are crucial for road planning.
The proposed graph modeling and GNN can capture such topological relationships, enabling decent location selection policies.
Table \ref{tab::overall} illustrates the performance of our method with and without graph modeling, \textit{i.e.}, DRL-GNN and DRL-MLP.
Specifically, it is easier for our graph model to perceive the currently disconnected regions, and layout corresponding road segments to connect them, leading to earlier universal connectivity in all 4 slums.
The graph model can also capture the neighborhood information on travel distance and construction costs, leading to a more economical policy to reduce travel distance.
As shown in Table \ref{tab::overall}, DRL-GNN outperforms DRL-MLP in AD and SC for 4 and 3 slums, respectively.
Furthermore, the graph modeling can also improve sample efficiency and makes our model converge faster (see Section \ref{app::graph_model} of the appendix).

\vspace{-3px}
\subsubsection*{\textbf{Topological Features}}
We investigate the role that the designed features for nodes, edges, and faces play in our model.
We first obtain a well-trained model, then remove different features, \textit{i.e.}, setting the feature values as 0, and evaluate its performance.
Figure \ref{fig::input_features} demonstrate the performance of removing three features (F1: \textit{Centrality}, F2: \textit{Road}, F3: \textit{Straightness}) compared with using all features.
We can observe that feature Straightness brings the largest performance deterioration in travel distance, with 22.0\% and 49.0\% increases in Harare and Cape Town, respectively.
This result is reasonable since \textit{Straightness} is the ratio of road network distance to the Euclidean distance, which directly indicates long detours in the slum, thus this feature is critical to travel distance.
In addition, feature \textit{Road} also plays an important role, and removing it leads to a 20.2\% and 13.0\% increase in construction cost for the two slums, respectively.
Results of removing other features can be found in Section \ref{app::input_features} of the appendix.
Our designed rich features describe the topological information of the slum, which is critical when selecting locations for new road segments.

\vspace{-3px}
\subsubsection*{\textbf{Topology-aware Message Passing}}
In the proposed GNN model, we design various propagation messages to edges from different sources, including nodes, faces, and edges themselves.
In this section, we study the effect of different propagation messages.
Specifically, we design multiple variants of our GNN model, each of which blocks one single propagation message.
We train these models and evaluate their road planning performance, as shown in Figure \ref{fig::message_passing}.
We can find that deleting any propagation flow leads to the loss of topological information, and brings about a deterioration in performance.
Specifically, deleting Node2Edge propagation makes travel distance worse by 10.2\% in Harare and construction cost worse by 8.1\% in Cape Town; deleting Edge Self-propagation increases construction cost by 14.8\% in Cape Town; and deleting Face2Edge propagation leads to a 3.0\% increase in travel distance in Harare.
The above results confirm the necessity of topology-aware message passing, which gathers diverse topological information to edges and makes our edge-centric GNN learn meaningful edge representations, enabling decent edge selection policies.

\begin{figure}[t]
    \centering
    \includegraphics[width=0.99\linewidth]{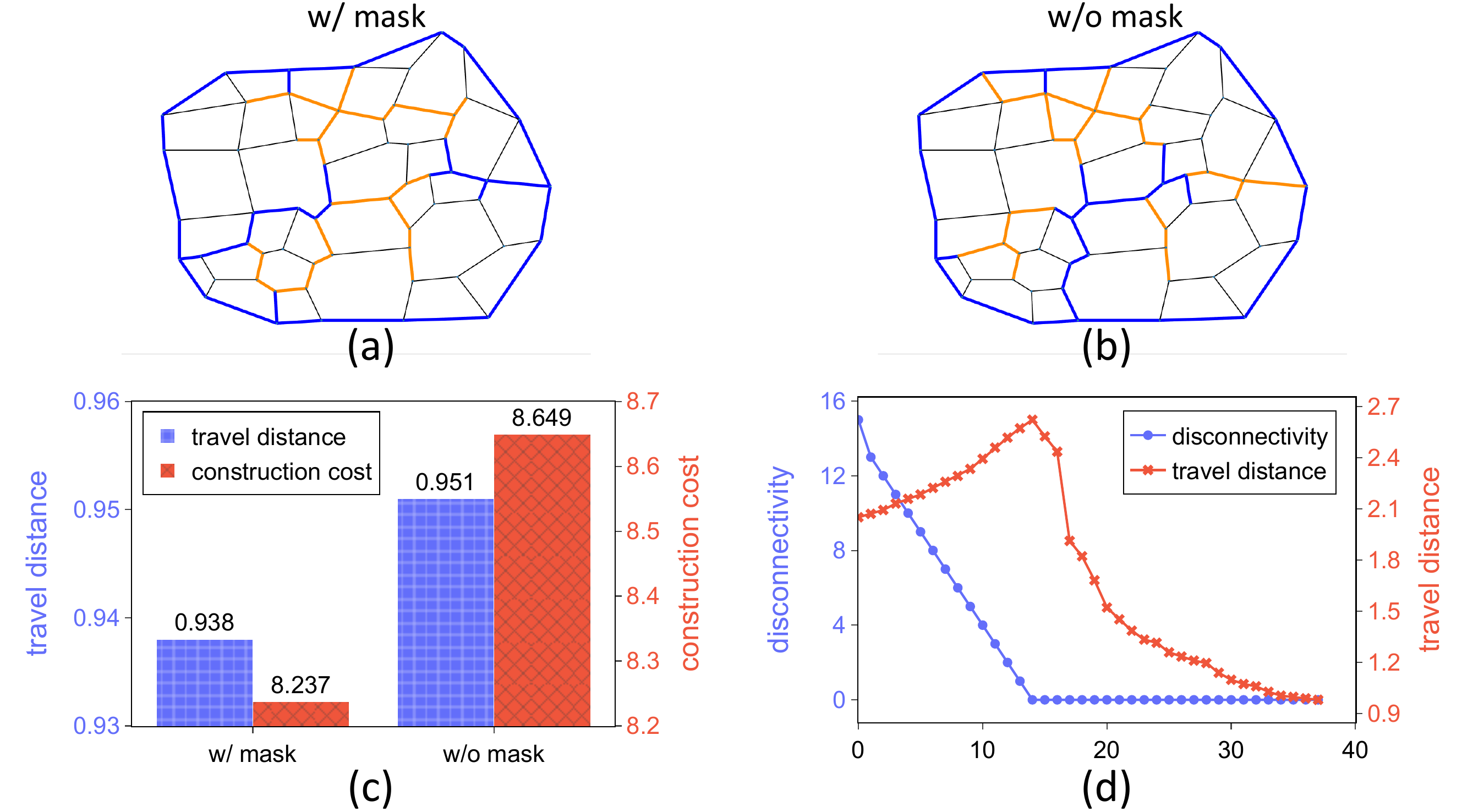}
    \vspace{-5px}
    \caption{
    The generated road plans for the slum in Cape Town, ZAF, of (a) DRL-GNN with action mask (b) DRL-GNN without action mask.
    (c) Travel distance and construction cost performance of the the two road plans.
    (d) The number of remaining disconnected places and travel distance at each step of our DRL-GNN model.
    Best viewed in color.
    }
    \vspace{-15px}
    \label{fig::stage_mask}
\end{figure}

\subsubsection*{\textbf{Masked Policy Optimization}}
We design a two-stage masked policy optimization method, to make our model first achieve universal connectivity and then reduce travel distance, at minimal costs in both stages.
In this section, we study the effect of the action mask, and Figure \ref{fig::stage_mask}(a-c) shows the generated road plans and the corresponding performance of our model with and without the action mask.
We can observe that adding the action mask can guide our model to discover disconnected places, and achieves lower travel distance (-1.4\%) with lower cost (-4.8\%) than without the mask.
Particularly, as in Table \ref{tab::overall}, even for the simple random model, adding the action mask can make it achieve universal connectivity faster than HS-MC for 3 slums.
Meanwhile, we visualize the road planning metrics at each step in Figure \ref{fig::stage_mask}(d).
We can see that in stage \uppercase\expandafter{\romannumeral1}, the number of remaining disconnected places decreases rapidly.
The travel distance increases slightly since newly connected places tend to have long travel distances to other places.
In stage \uppercase\expandafter{\romannumeral2}, the model adds road segments to places that suffer from long detours, significantly reducing the travel distance.
Furthermore, the masked policy optimization design avoids actions of low quality, which helps our model converge faster (see Section \ref{app::action_mask} of the appendix).
The results verify the effectiveness of the action mask and two-stage design in optimizing multiple objectives of road planning for slums.

We leave hyper-parameter study in Section \ref{app::hyper} of the appendix.
In addition, our proposed method is a flexible framework, making it easy to integrate with advanced network structures such as graph transformer~\cite{yun2019graph} (see Section \ref{app::graph_transformer} of the appendix).

\subsection{Analysis on Transferability}

\begin{figure}[t]
    \centering
    \includegraphics[width=0.99\linewidth]{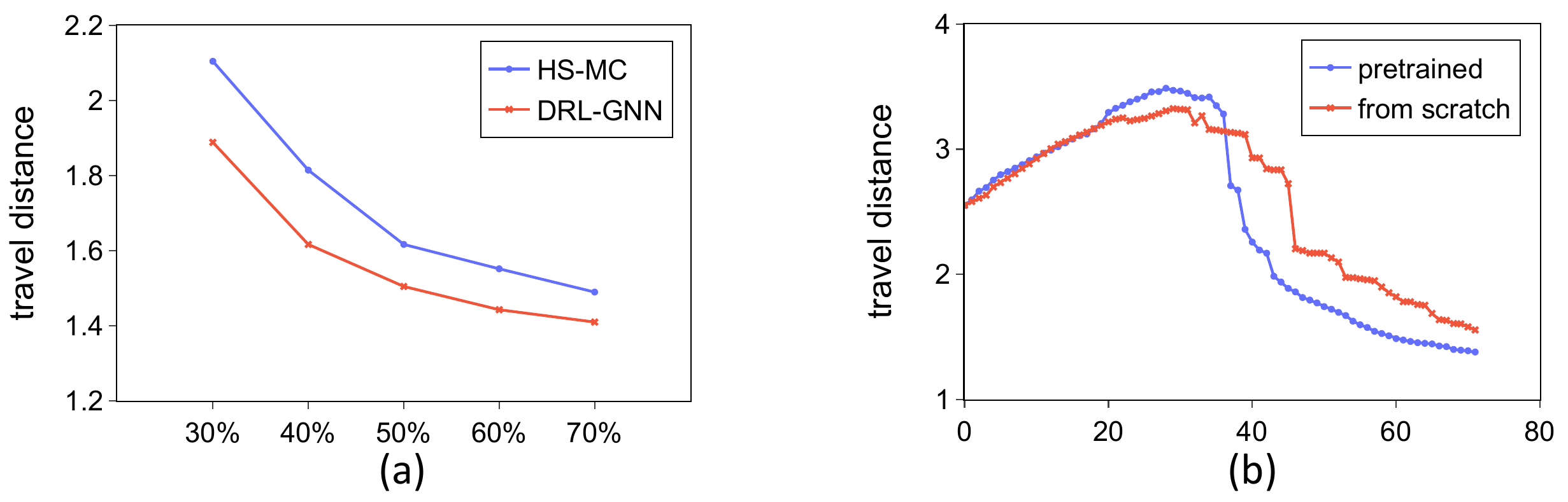}
    \vspace{-5px}
    \caption{
    (a) The travel distance of HSMC and DRL-GNN under different planning budgets.
    (b) The travel distance at each step of our DRL-GNN model.
    Best viewed in color.
    }
    \vspace{-15px}
    \label{fig::transfer}
\end{figure}

It is beneficial for a road planning model to generalize across different scenarios.
On the one hand, the planning budgets may vary.
We set the budget as 50\% of candidate segments for training, and directly evaluate our model under different budgets.
Figure \ref{fig::transfer}(a) shows that DRL-GNN outperforms HS-MC under all different budgets, with more significant improvements under tight budgets, \textit{e.g.}, 10.3\% travel distance reduction under 30\% budgets, two times larger than 70\% budgets.
On the other hand, we study the transferability across different slums.
We obtain a pretrained model on a small slum (Harare, ZWE), and finetune it on a large slum (Cape Town, ZAF).
We compare the pretrained model with a model that is trained from scratch.
Figure \ref{fig::transfer}(b) demonstrates the travel distance at each step for the large slum, where the pretrained model is consistently better than the model trained from scratch in stage \uppercase\expandafter{\romannumeral2}.
The above results verify that our model can learn universal road planning skills and successfully transfer them to scenarios of different budgets or different slums, which is crucial for practical slum upgrading.

\section{Related Work}

\noindent\textbf{Deep Reinforcement Learning for Planning.}
With the development of deep learning~\cite{lecun2015deep}, utilizing deep neural networks (DNN) to achieve function approximation in reinforcement learning becomes the new state-of-the-art.
Since the proposal of DQN~\cite{mnih2013playing,mnih2015human}, DRL methods have achieved great success in complex planning tasks, such as the game of Go~\cite{silver2016mastering,silver2017mastering}, chemical synthesis~\cite{segler2018planning}, chip design~\cite{amini2022generalizable,roy2021prefixrl,mirhoseini2021graph}, VRP~\cite{nazari2018reinforcement,duan2020efficiently,zong2022rbg}, and solving mathematical problems~\cite{fawzi2022discovering}.
Planning tasks usually have a huge action space, which can be effectively reduced by predicting rewards with a value network and sampling actions via a policy network~\cite{konda1999actor,silver2016mastering,haarnoja2018soft,mnih2016asynchronous}.
Recently, several works~\cite{fan2020finding,meirom2021controlling,wang2018nervenet,lei2020reinforcement} adopt GNN as policy and value networks to solve planning tasks on the graph~\cite{fan2020finding,meirom2021controlling}.
For example, Fan \textit{et al.}~\cite{fan2020finding} combine GNN with DQN to detect key nodes in complex networks.
Meirom \textit{et al.}~\cite{meirom2021controlling} utilize GNN as a state encoder for DRL to solve the tasks of epidemic control and targeted marketing.
In addition, GNN is leveraged to learn representations for road networks~\cite{jepsen2019graph,wu2020learning,xue2022quantifying,wang2020representation,hu2019stochastic,mao2022jointly,derrow2021eta}, and support downstream tasks like homogeneity analysis~\cite{xue2022quantifying} and traffic prediction~\cite{wu2020learning,derrow2021eta}.
However, they only study tasks on existing built roads, which is quite different from the task of planning new roads.
Meanwhile, there have been some works utilizing DRL or generative models to accomplish city configuration and urban planning~\cite{wang2022human,wang2021reinforced,wang2023automated,wang2020reimagining,wang2021deep,fang2022incorporating,kempinska2019modelling,geng2021deep}, but they ignore the slums in cities which is an important issue regarding billions of population.
In this work, we make the first attempt to plan new roads for slum upgrading with DRL and GNN.

\noindent\textbf{Road Planning for Slums.}
Given the large number of slum dwellers and the economic costs, upgrading slums in situ has become the primary strategy of urbanization, rather than relocating the population to cities.
One primary goal of slum upgrading is to provide service access to every place in a slum by building more roads.
The re-blocking method~\cite{mitlin2012urban,habitat2012streets,weru2004community,boonyabancha2005baan,patel2012knowledge,mitlin2012urban} is widely adopted in practice, which reconfigures the space and adds road segments, to make each place connected to the road system.
With more streets constructed, re-blocking has been shown to significantly reduce the cost of service provision for slums~\cite{abiko2007basic,flood2004cost}.
However, it is not a computational method and requires negotiation with multiple stakeholders, so it is slow and case-by-case.
A recent paper by Brelsford \textit{et al}~\cite{brelsford2018toward} formulated road planning for slums as a constrained optimization problem, making it computationally solvable.
Specifically, they proposed a heuristic search approach, adding one path at a time to the least connected place, with the help of Monte Carlo sampling.
Considering the huge solution space of this problem, it is difficult for heuristic methods to achieve optimal road planning performance.
Different from heuristic search, in this work, we leverage the powerful DRL algorithm to search for optimal road plans in a data-driven way, improving accessibility at minimal costs.

\section{Conclusion}

In this paper, we investigate the problem of road planning for slums, a critical but little-studied issue in sustainable urban development.
We formulate it as a sequential decision-making problem with a generic graph model, and propose a novel graph neural network to select locations for new road segments.
The model is optimized to improve accessibility and reduce the travel distance of slum dwellers at minimal construction costs.
We demonstrate that planning roads for slums through deep reinforcement learning is viable, effective, and can be migrated to real-world, large-scale scenarios.
As for future work, we plan to develop a pre-trained model on a large amount of slum data to enable fast inference of road plans, which is beneficial for practical deployment.

\begin{acks}
We sincerely thank all reviewers for their constructive feedback.
We are grateful to Yuming Lin for valuable discussions.
This work is supported in part by National Key Research and Development Program of China under grant 2022ZD0116402, National Natural Science Foundation of China under grant U22B2057, U21B2036, 62171260.
This work is also supported in part by Beijing National Research Center for Information Science and Technology (BNRist).
\end{acks}

\bibliographystyle{ACM-Reference-Format}
\balance
\bibliography{references}


\begin{thebibliography}{74}


\ifx \showCODEN    \undefined \def \showCODEN     #1{\unskip}     \fi
\ifx \showDOI      \undefined \def \showDOI       #1{#1}\fi
\ifx \showISBNx    \undefined \def \showISBNx     #1{\unskip}     \fi
\ifx \showISBNxiii \undefined \def \showISBNxiii  #1{\unskip}     \fi
\ifx \showISSN     \undefined \def \showISSN      #1{\unskip}     \fi
\ifx \showLCCN     \undefined \def \showLCCN      #1{\unskip}     \fi
\ifx \shownote     \undefined \def \shownote      #1{#1}          \fi
\ifx \showarticletitle \undefined \def \showarticletitle #1{#1}   \fi
\ifx \showURL      \undefined \def \showURL       {\relax}        \fi
\providecommand\bibfield[2]{#2}
\providecommand\bibinfo[2]{#2}
\providecommand\natexlab[1]{#1}
\providecommand\showeprint[2][]{arXiv:#2}

\bibitem[Abiko et~al\mbox{.}(2007)]%
        {abiko2007basic}
\bibfield{author}{\bibinfo{person}{Alex Abiko}, \bibinfo{person}{Luiz~Reynaldo
  de Azevedo~Cardoso}, \bibinfo{person}{Ricardo Rinaldelli}, {and}
  \bibinfo{person}{Heitor Cesar~Riogi Haga}.} \bibinfo{year}{2007}\natexlab{}.
\newblock \showarticletitle{Basic costs of slum upgrading in Brazil}.
\newblock \bibinfo{journal}{\emph{Global Urban Development Magazine}}
  \bibinfo{volume}{3}, \bibinfo{number}{1} (\bibinfo{year}{2007}),
  \bibinfo{pages}{121--131}.
\newblock


\bibitem[Alexander et~al\mbox{.}(2019)]%
        {alexander2019city}
\bibfield{author}{\bibinfo{person}{Christopher Alexander} {et~al\mbox{.}}}
  \bibinfo{year}{2019}\natexlab{}.
\newblock \bibinfo{title}{A city is not a tree}.
\newblock
\newblock


\bibitem[Amini et~al\mbox{.}(2022)]%
        {amini2022generalizable}
\bibfield{author}{\bibinfo{person}{Mohammad Amini}, \bibinfo{person}{Zhanguang
  Zhang}, \bibinfo{person}{Surya Penmetsa}, \bibinfo{person}{Yingxue Zhang},
  \bibinfo{person}{Jianye Hao}, {and} \bibinfo{person}{Wulong Liu}.}
  \bibinfo{year}{2022}\natexlab{}.
\newblock \showarticletitle{Generalizable Floorplanner through Corner Block
  List Representation and Hypergraph Embedding}. In
  \bibinfo{booktitle}{\emph{Proceedings of the 28th ACM SIGKDD Conference on
  Knowledge Discovery and Data Mining}}. \bibinfo{pages}{2692--2702}.
\newblock


\bibitem[Andavarapu et~al\mbox{.}(2013)]%
        {andavarapu2013evolution}
\bibfield{author}{\bibinfo{person}{Deepika Andavarapu},
  \bibinfo{person}{David~J Edelman}, {et~al\mbox{.}}}
  \bibinfo{year}{2013}\natexlab{}.
\newblock \showarticletitle{Evolution of slum redevelopment policy}.
\newblock \bibinfo{journal}{\emph{Current urban studies}} \bibinfo{volume}{1},
  \bibinfo{number}{04} (\bibinfo{year}{2013}), \bibinfo{pages}{185}.
\newblock


\bibitem[Barth{\'e}lemy and Flammini(2008)]%
        {barthelemy2008modeling}
\bibfield{author}{\bibinfo{person}{Marc Barth{\'e}lemy} {and}
  \bibinfo{person}{Alessandro Flammini}.} \bibinfo{year}{2008}\natexlab{}.
\newblock \showarticletitle{Modeling urban street patterns}.
\newblock \bibinfo{journal}{\emph{Physical review letters}}
  \bibinfo{volume}{100}, \bibinfo{number}{13} (\bibinfo{year}{2008}),
  \bibinfo{pages}{138702}.
\newblock


\bibitem[Binder et~al\mbox{.}(1993)]%
        {binder1993monte}
\bibfield{author}{\bibinfo{person}{Kurt Binder}, \bibinfo{person}{Dieter
  Heermann}, \bibinfo{person}{Lyle Roelofs}, \bibinfo{person}{A~John
  Mallinckrodt}, {and} \bibinfo{person}{Susan McKay}.}
  \bibinfo{year}{1993}\natexlab{}.
\newblock \showarticletitle{Monte Carlo simulation in statistical physics}.
\newblock \bibinfo{journal}{\emph{Computers in Physics}} \bibinfo{volume}{7},
  \bibinfo{number}{2} (\bibinfo{year}{1993}), \bibinfo{pages}{156--157}.
\newblock


\bibitem[Bonacich(1987)]%
        {bonacich1987power}
\bibfield{author}{\bibinfo{person}{Phillip Bonacich}.}
  \bibinfo{year}{1987}\natexlab{}.
\newblock \showarticletitle{Power and centrality: A family of measures}.
\newblock \bibinfo{journal}{\emph{American journal of sociology}}
  \bibinfo{volume}{92}, \bibinfo{number}{5} (\bibinfo{year}{1987}),
  \bibinfo{pages}{1170--1182}.
\newblock


\bibitem[Boonyabancha(2005)]%
        {boonyabancha2005baan}
\bibfield{author}{\bibinfo{person}{Somsook Boonyabancha}.}
  \bibinfo{year}{2005}\natexlab{}.
\newblock \showarticletitle{Baan Mankong: Going to scale with “slum” and
  squatter upgrading in Thailand}.
\newblock \bibinfo{journal}{\emph{Environment and Urbanization}}
  \bibinfo{volume}{17}, \bibinfo{number}{1} (\bibinfo{year}{2005}),
  \bibinfo{pages}{21--46}.
\newblock


\bibitem[Brelsford et~al\mbox{.}(2018)]%
        {brelsford2018toward}
\bibfield{author}{\bibinfo{person}{Christa Brelsford}, \bibinfo{person}{Taylor
  Martin}, \bibinfo{person}{Joe Hand}, {and} \bibinfo{person}{Lu{\'\i}s~MA
  Bettencourt}.} \bibinfo{year}{2018}\natexlab{}.
\newblock \showarticletitle{Toward cities without slums: Topology and the
  spatial evolution of neighborhoods}.
\newblock \bibinfo{journal}{\emph{Science advances}} \bibinfo{volume}{4},
  \bibinfo{number}{8} (\bibinfo{year}{2018}), \bibinfo{pages}{eaar4644}.
\newblock


\bibitem[Chen et~al\mbox{.}(2020)]%
        {chen2020measuring}
\bibfield{author}{\bibinfo{person}{Deli Chen}, \bibinfo{person}{Yankai Lin},
  \bibinfo{person}{Wei Li}, \bibinfo{person}{Peng Li}, \bibinfo{person}{Jie
  Zhou}, {and} \bibinfo{person}{Xu Sun}.} \bibinfo{year}{2020}\natexlab{}.
\newblock \showarticletitle{Measuring and relieving the over-smoothing problem
  for graph neural networks from the topological view}. In
  \bibinfo{booktitle}{\emph{Proceedings of the AAAI conference on artificial
  intelligence}}, Vol.~\bibinfo{volume}{34}. \bibinfo{pages}{3438--3445}.
\newblock


\bibitem[Corburn and Karanja(2014)]%
        {corburn2014informal}
\bibfield{author}{\bibinfo{person}{Jason Corburn} {and} \bibinfo{person}{Irene
  Karanja}.} \bibinfo{year}{2014}\natexlab{}.
\newblock \showarticletitle{Informal settlements and a relational view of
  health in Nairobi, Kenya: sanitation, gender and dignity}.
\newblock \bibinfo{journal}{\emph{Health promotion international}}
  \bibinfo{volume}{31}, \bibinfo{number}{2} (\bibinfo{year}{2014}),
  \bibinfo{pages}{258--269}.
\newblock


\bibitem[Derrow-Pinion et~al\mbox{.}(2021)]%
        {derrow2021eta}
\bibfield{author}{\bibinfo{person}{Austin Derrow-Pinion},
  \bibinfo{person}{Jennifer She}, \bibinfo{person}{David Wong},
  \bibinfo{person}{Oliver Lange}, \bibinfo{person}{Todd Hester},
  \bibinfo{person}{Luis Perez}, \bibinfo{person}{Marc Nunkesser},
  \bibinfo{person}{Seongjae Lee}, \bibinfo{person}{Xueying Guo},
  \bibinfo{person}{Brett Wiltshire}, {et~al\mbox{.}}}
  \bibinfo{year}{2021}\natexlab{}.
\newblock \showarticletitle{Eta prediction with graph neural networks in google
  maps}. In \bibinfo{booktitle}{\emph{Proceedings of the 30th ACM International
  Conference on Information \& Knowledge Management}}.
  \bibinfo{pages}{3767--3776}.
\newblock


\bibitem[Duan et~al\mbox{.}(2020)]%
        {duan2020efficiently}
\bibfield{author}{\bibinfo{person}{Lu Duan}, \bibinfo{person}{Yang Zhan},
  \bibinfo{person}{Haoyuan Hu}, \bibinfo{person}{Yu Gong},
  \bibinfo{person}{Jiangwen Wei}, \bibinfo{person}{Xiaodong Zhang}, {and}
  \bibinfo{person}{Yinghui Xu}.} \bibinfo{year}{2020}\natexlab{}.
\newblock \showarticletitle{Efficiently solving the practical vehicle routing
  problem: A novel joint learning approach}. In
  \bibinfo{booktitle}{\emph{Proceedings of the 26th ACM SIGKDD international
  conference on knowledge discovery \& data mining}}.
  \bibinfo{pages}{3054--3063}.
\newblock


\bibitem[Fan et~al\mbox{.}(2020)]%
        {fan2020finding}
\bibfield{author}{\bibinfo{person}{Changjun Fan}, \bibinfo{person}{Li Zeng},
  \bibinfo{person}{Yizhou Sun}, {and} \bibinfo{person}{Yang-Yu Liu}.}
  \bibinfo{year}{2020}\natexlab{}.
\newblock \showarticletitle{Finding key players in complex networks through
  deep reinforcement learning}.
\newblock \bibinfo{journal}{\emph{Nature machine intelligence}}
  \bibinfo{volume}{2}, \bibinfo{number}{6} (\bibinfo{year}{2020}),
  \bibinfo{pages}{317--324}.
\newblock


\bibitem[Fang et~al\mbox{.}(2022)]%
        {fang2022incorporating}
\bibfield{author}{\bibinfo{person}{Zhou Fang}, \bibinfo{person}{Ying Jin},
  {and} \bibinfo{person}{Tianren Yang}.} \bibinfo{year}{2022}\natexlab{}.
\newblock \showarticletitle{Incorporating planning intelligence into deep
  learning: A planning support tool for street network design}.
\newblock \bibinfo{journal}{\emph{Journal of Urban Technology}}
  \bibinfo{volume}{29}, \bibinfo{number}{2} (\bibinfo{year}{2022}),
  \bibinfo{pages}{99--114}.
\newblock


\bibitem[Farahani et~al\mbox{.}(2013)]%
        {farahani2013review}
\bibfield{author}{\bibinfo{person}{Reza~Zanjirani Farahani},
  \bibinfo{person}{Elnaz Miandoabchi}, \bibinfo{person}{Wai~Yuen Szeto}, {and}
  \bibinfo{person}{Hannaneh Rashidi}.} \bibinfo{year}{2013}\natexlab{}.
\newblock \showarticletitle{A review of urban transportation network design
  problems}.
\newblock \bibinfo{journal}{\emph{European journal of operational research}}
  \bibinfo{volume}{229}, \bibinfo{number}{2} (\bibinfo{year}{2013}),
  \bibinfo{pages}{281--302}.
\newblock


\bibitem[Fawzi et~al\mbox{.}(2022)]%
        {fawzi2022discovering}
\bibfield{author}{\bibinfo{person}{Alhussein Fawzi}, \bibinfo{person}{Matej
  Balog}, \bibinfo{person}{Aja Huang}, \bibinfo{person}{Thomas Hubert},
  \bibinfo{person}{Bernardino Romera-Paredes}, \bibinfo{person}{Mohammadamin
  Barekatain}, \bibinfo{person}{Alexander Novikov},
  \bibinfo{person}{Francisco~J R~Ruiz}, \bibinfo{person}{Julian Schrittwieser},
  \bibinfo{person}{Grzegorz Swirszcz}, {et~al\mbox{.}}}
  \bibinfo{year}{2022}\natexlab{}.
\newblock \showarticletitle{Discovering faster matrix multiplication algorithms
  with reinforcement learning}.
\newblock \bibinfo{journal}{\emph{Nature}} \bibinfo{volume}{610},
  \bibinfo{number}{7930} (\bibinfo{year}{2022}), \bibinfo{pages}{47--53}.
\newblock


\bibitem[Flood(2004)]%
        {flood2004cost}
\bibfield{author}{\bibinfo{person}{Joe Flood}.}
  \bibinfo{year}{2004}\natexlab{}.
\newblock \showarticletitle{Cost Estimate for Millennium Development Goal 7,
  Target 11 on Slums, background report for UN Millennium Project Task Force on
  Improving the Lives of Slum Dwellers and UN-HABITAT}.
\newblock \bibinfo{journal}{\emph{Urban Resources, Elsternwick, Australia}}
  (\bibinfo{year}{2004}).
\newblock


\bibitem[Freeman(1977)]%
        {freeman1977set}
\bibfield{author}{\bibinfo{person}{Linton~C Freeman}.}
  \bibinfo{year}{1977}\natexlab{}.
\newblock \showarticletitle{A set of measures of centrality based on
  betweenness}.
\newblock \bibinfo{journal}{\emph{Sociometry}} (\bibinfo{year}{1977}),
  \bibinfo{pages}{35--41}.
\newblock


\bibitem[Freeman et~al\mbox{.}(2002)]%
        {freeman2002centrality}
\bibfield{author}{\bibinfo{person}{Linton~C Freeman} {et~al\mbox{.}}}
  \bibinfo{year}{2002}\natexlab{}.
\newblock \showarticletitle{Centrality in social networks: Conceptual
  clarification}.
\newblock \bibinfo{journal}{\emph{Social network: critical concepts in
  sociology. Londres: Routledge}}  \bibinfo{volume}{1} (\bibinfo{year}{2002}),
  \bibinfo{pages}{238--263}.
\newblock


\bibitem[Gad(2021)]%
        {gad2021pygad}
\bibfield{author}{\bibinfo{person}{Ahmed~Fawzy Gad}.}
  \bibinfo{year}{2021}\natexlab{}.
\newblock \bibinfo{title}{PyGAD: An Intuitive Genetic Algorithm Python
  Library}.
\newblock
\newblock
\showeprint[arxiv]{2106.06158}~[cs.NE]


\bibitem[Geng et~al\mbox{.}(2021)]%
        {geng2021deep}
\bibfield{author}{\bibinfo{person}{Yuanzhe Geng}, \bibinfo{person}{Erwu Liu},
  \bibinfo{person}{Rui Wang}, \bibinfo{person}{Yiming Liu},
  \bibinfo{person}{Weixiong Rao}, \bibinfo{person}{Shaojun Feng},
  \bibinfo{person}{Zhao Dong}, \bibinfo{person}{Zhiren Fu}, {and}
  \bibinfo{person}{Yanfen Chen}.} \bibinfo{year}{2021}\natexlab{}.
\newblock \showarticletitle{Deep reinforcement learning based dynamic route
  planning for minimizing travel time}. In \bibinfo{booktitle}{\emph{2021 IEEE
  International Conference on Communications Workshops (ICC Workshops)}}. IEEE,
  \bibinfo{pages}{1--6}.
\newblock


\bibitem[Goodfellow et~al\mbox{.}(2020)]%
        {goodfellow2020generative}
\bibfield{author}{\bibinfo{person}{Ian Goodfellow}, \bibinfo{person}{Jean
  Pouget-Abadie}, \bibinfo{person}{Mehdi Mirza}, \bibinfo{person}{Bing Xu},
  \bibinfo{person}{David Warde-Farley}, \bibinfo{person}{Sherjil Ozair},
  \bibinfo{person}{Aaron Courville}, {and} \bibinfo{person}{Yoshua Bengio}.}
  \bibinfo{year}{2020}\natexlab{}.
\newblock \showarticletitle{Generative adversarial networks}.
\newblock \bibinfo{journal}{\emph{Commun. ACM}} \bibinfo{volume}{63},
  \bibinfo{number}{11} (\bibinfo{year}{2020}), \bibinfo{pages}{139--144}.
\newblock


\bibitem[Haarnoja et~al\mbox{.}(2018)]%
        {haarnoja2018soft}
\bibfield{author}{\bibinfo{person}{Tuomas Haarnoja}, \bibinfo{person}{Aurick
  Zhou}, \bibinfo{person}{Pieter Abbeel}, {and} \bibinfo{person}{Sergey
  Levine}.} \bibinfo{year}{2018}\natexlab{}.
\newblock \showarticletitle{Soft actor-critic: Off-policy maximum entropy deep
  reinforcement learning with a stochastic actor}. In
  \bibinfo{booktitle}{\emph{International conference on machine learning}}.
  PMLR, \bibinfo{pages}{1861--1870}.
\newblock


\bibitem[Habitat(2012)]%
        {habitat2012streets}
\bibfield{author}{\bibinfo{person}{UN Habitat}.}
  \bibinfo{year}{2012}\natexlab{}.
\newblock \showarticletitle{Streets as tools for urban transformation in slums:
  a street-led approach to citywide slum upgrading}.
\newblock   \bibinfo{volume}{17} (\bibinfo{year}{2012}), \bibinfo{pages}{2016}.
\newblock


\bibitem[Hamilton et~al\mbox{.}(2017)]%
        {hamilton2017inductive}
\bibfield{author}{\bibinfo{person}{Will Hamilton}, \bibinfo{person}{Zhitao
  Ying}, {and} \bibinfo{person}{Jure Leskovec}.}
  \bibinfo{year}{2017}\natexlab{}.
\newblock \showarticletitle{Inductive representation learning on large graphs}.
\newblock \bibinfo{journal}{\emph{Advances in neural information processing
  systems}}  \bibinfo{volume}{30} (\bibinfo{year}{2017}).
\newblock


\bibitem[He et~al\mbox{.}(2020)]%
        {he2020lightgcn}
\bibfield{author}{\bibinfo{person}{Xiangnan He}, \bibinfo{person}{Kuan Deng},
  \bibinfo{person}{Xiang Wang}, \bibinfo{person}{Yan Li},
  \bibinfo{person}{Yongdong Zhang}, {and} \bibinfo{person}{Meng Wang}.}
  \bibinfo{year}{2020}\natexlab{}.
\newblock \showarticletitle{Lightgcn: Simplifying and powering graph
  convolution network for recommendation}. In
  \bibinfo{booktitle}{\emph{Proceedings of the 43rd International ACM SIGIR
  conference on research and development in Information Retrieval}}.
  \bibinfo{pages}{639--648}.
\newblock


\bibitem[Hu et~al\mbox{.}(2019)]%
        {hu2019stochastic}
\bibfield{author}{\bibinfo{person}{Jilin Hu}, \bibinfo{person}{Chenjuan Guo},
  \bibinfo{person}{Bin Yang}, {and} \bibinfo{person}{Christian~S Jensen}.}
  \bibinfo{year}{2019}\natexlab{}.
\newblock \showarticletitle{Stochastic weight completion for road networks
  using graph convolutional networks}. In \bibinfo{booktitle}{\emph{2019 IEEE
  35th international conference on data engineering (ICDE)}}. IEEE,
  \bibinfo{pages}{1274--1285}.
\newblock


\bibitem[Jepsen et~al\mbox{.}(2019)]%
        {jepsen2019graph}
\bibfield{author}{\bibinfo{person}{Tobias~Skovgaard Jepsen},
  \bibinfo{person}{Christian~S Jensen}, {and} \bibinfo{person}{Thomas~Dyhre
  Nielsen}.} \bibinfo{year}{2019}\natexlab{}.
\newblock \showarticletitle{Graph convolutional networks for road networks}. In
  \bibinfo{booktitle}{\emph{Proceedings of the 27th ACM SIGSPATIAL
  international conference on advances in geographic information systems}}.
  \bibinfo{pages}{460--463}.
\newblock


\bibitem[Kawaguchi et~al\mbox{.}(2017)]%
        {kawaguchi2017generalization}
\bibfield{author}{\bibinfo{person}{Kenji Kawaguchi},
  \bibinfo{person}{Leslie~Pack Kaelbling}, {and} \bibinfo{person}{Yoshua
  Bengio}.} \bibinfo{year}{2017}\natexlab{}.
\newblock \showarticletitle{Generalization in deep learning}.
\newblock \bibinfo{journal}{\emph{arXiv preprint arXiv:1710.05468}}
  (\bibinfo{year}{2017}).
\newblock


\bibitem[Kempinska and Murcio(2019)]%
        {kempinska2019modelling}
\bibfield{author}{\bibinfo{person}{Kira Kempinska} {and}
  \bibinfo{person}{Roberto Murcio}.} \bibinfo{year}{2019}\natexlab{}.
\newblock \showarticletitle{Modelling urban networks using Variational
  Autoencoders}.
\newblock \bibinfo{journal}{\emph{Applied Network Science}}
  \bibinfo{volume}{4}, \bibinfo{number}{1} (\bibinfo{year}{2019}),
  \bibinfo{pages}{1--11}.
\newblock


\bibitem[Kingma and Welling(2013)]%
        {kingma2013auto}
\bibfield{author}{\bibinfo{person}{Diederik~P Kingma} {and}
  \bibinfo{person}{Max Welling}.} \bibinfo{year}{2013}\natexlab{}.
\newblock \showarticletitle{Auto-encoding variational bayes}.
\newblock \bibinfo{journal}{\emph{arXiv preprint arXiv:1312.6114}}
  (\bibinfo{year}{2013}).
\newblock


\bibitem[Kipf and Welling(2016)]%
        {kipf2016semi}
\bibfield{author}{\bibinfo{person}{Thomas~N Kipf} {and} \bibinfo{person}{Max
  Welling}.} \bibinfo{year}{2016}\natexlab{}.
\newblock \showarticletitle{Semi-supervised classification with graph
  convolutional networks}.
\newblock \bibinfo{journal}{\emph{arXiv preprint arXiv:1609.02907}}
  (\bibinfo{year}{2016}).
\newblock


\bibitem[Konda and Tsitsiklis(1999)]%
        {konda1999actor}
\bibfield{author}{\bibinfo{person}{Vijay Konda} {and} \bibinfo{person}{John
  Tsitsiklis}.} \bibinfo{year}{1999}\natexlab{}.
\newblock \showarticletitle{Actor-critic algorithms}.
\newblock \bibinfo{journal}{\emph{Advances in neural information processing
  systems}}  \bibinfo{volume}{12} (\bibinfo{year}{1999}).
\newblock


\bibitem[Kruskal(1956)]%
        {kruskal1956shortest}
\bibfield{author}{\bibinfo{person}{Joseph~B Kruskal}.}
  \bibinfo{year}{1956}\natexlab{}.
\newblock \showarticletitle{On the shortest spanning subtree of a graph and the
  traveling salesman problem}.
\newblock \bibinfo{journal}{\emph{Proceedings of the American Mathematical
  society}} \bibinfo{volume}{7}, \bibinfo{number}{1} (\bibinfo{year}{1956}),
  \bibinfo{pages}{48--50}.
\newblock


\bibitem[LeCun et~al\mbox{.}(2015)]%
        {lecun2015deep}
\bibfield{author}{\bibinfo{person}{Yann LeCun}, \bibinfo{person}{Yoshua
  Bengio}, {and} \bibinfo{person}{Geoffrey Hinton}.}
  \bibinfo{year}{2015}\natexlab{}.
\newblock \showarticletitle{Deep learning}.
\newblock \bibinfo{journal}{\emph{nature}} \bibinfo{volume}{521},
  \bibinfo{number}{7553} (\bibinfo{year}{2015}), \bibinfo{pages}{436--444}.
\newblock


\bibitem[Lei et~al\mbox{.}(2020)]%
        {lei2020reinforcement}
\bibfield{author}{\bibinfo{person}{Yu Lei}, \bibinfo{person}{Hongbin Pei},
  \bibinfo{person}{Hanqi Yan}, {and} \bibinfo{person}{Wenjie Li}.}
  \bibinfo{year}{2020}\natexlab{}.
\newblock \showarticletitle{Reinforcement learning based recommendation with
  graph convolutional q-network}. In \bibinfo{booktitle}{\emph{Proceedings of
  the 43rd international ACM SIGIR conference on research and development in
  information retrieval}}. \bibinfo{pages}{1757--1760}.
\newblock


\bibitem[Mao et~al\mbox{.}(2022)]%
        {mao2022jointly}
\bibfield{author}{\bibinfo{person}{Zhenyu Mao}, \bibinfo{person}{Ziyue Li},
  \bibinfo{person}{Dedong Li}, \bibinfo{person}{Lei Bai}, {and}
  \bibinfo{person}{Rui Zhao}.} \bibinfo{year}{2022}\natexlab{}.
\newblock \showarticletitle{Jointly Contrastive Representation Learning on Road
  Network and Trajectory}. In \bibinfo{booktitle}{\emph{Proceedings of the 31st
  ACM International Conference on Information \& Knowledge Management}}.
  \bibinfo{pages}{1501--1510}.
\newblock


\bibitem[Meirom et~al\mbox{.}(2021)]%
        {meirom2021controlling}
\bibfield{author}{\bibinfo{person}{Eli Meirom}, \bibinfo{person}{Haggai Maron},
  \bibinfo{person}{Shie Mannor}, {and} \bibinfo{person}{Gal Chechik}.}
  \bibinfo{year}{2021}\natexlab{}.
\newblock \showarticletitle{Controlling graph dynamics with reinforcement
  learning and graph neural networks}. In
  \bibinfo{booktitle}{\emph{International Conference on Machine Learning}}.
  PMLR, \bibinfo{pages}{7565--7577}.
\newblock


\bibitem[Mirhoseini et~al\mbox{.}(2021)]%
        {mirhoseini2021graph}
\bibfield{author}{\bibinfo{person}{Azalia Mirhoseini}, \bibinfo{person}{Anna
  Goldie}, \bibinfo{person}{Mustafa Yazgan}, \bibinfo{person}{Joe~Wenjie
  Jiang}, \bibinfo{person}{Ebrahim Songhori}, \bibinfo{person}{Shen Wang},
  \bibinfo{person}{Young-Joon Lee}, \bibinfo{person}{Eric Johnson},
  \bibinfo{person}{Omkar Pathak}, \bibinfo{person}{Azade Nazi},
  {et~al\mbox{.}}} \bibinfo{year}{2021}\natexlab{}.
\newblock \showarticletitle{A graph placement methodology for fast chip
  design}.
\newblock \bibinfo{journal}{\emph{Nature}} \bibinfo{volume}{594},
  \bibinfo{number}{7862} (\bibinfo{year}{2021}), \bibinfo{pages}{207--212}.
\newblock


\bibitem[Mitlin and Satterthwaite(2012)]%
        {mitlin2012urban}
\bibfield{author}{\bibinfo{person}{Diana Mitlin} {and} \bibinfo{person}{David
  Satterthwaite}.} \bibinfo{year}{2012}\natexlab{}.
\newblock \bibinfo{booktitle}{\emph{Urban poverty in the global south: scale
  and nature}}.
\newblock \bibinfo{publisher}{Routledge}.
\newblock


\bibitem[Mnih et~al\mbox{.}(2016)]%
        {mnih2016asynchronous}
\bibfield{author}{\bibinfo{person}{Volodymyr Mnih},
  \bibinfo{person}{Adria~Puigdomenech Badia}, \bibinfo{person}{Mehdi Mirza},
  \bibinfo{person}{Alex Graves}, \bibinfo{person}{Timothy Lillicrap},
  \bibinfo{person}{Tim Harley}, \bibinfo{person}{David Silver}, {and}
  \bibinfo{person}{Koray Kavukcuoglu}.} \bibinfo{year}{2016}\natexlab{}.
\newblock \showarticletitle{Asynchronous methods for deep reinforcement
  learning}. In \bibinfo{booktitle}{\emph{International conference on machine
  learning}}. PMLR, \bibinfo{pages}{1928--1937}.
\newblock


\bibitem[Mnih et~al\mbox{.}(2013)]%
        {mnih2013playing}
\bibfield{author}{\bibinfo{person}{Volodymyr Mnih}, \bibinfo{person}{Koray
  Kavukcuoglu}, \bibinfo{person}{David Silver}, \bibinfo{person}{Alex Graves},
  \bibinfo{person}{Ioannis Antonoglou}, \bibinfo{person}{Daan Wierstra}, {and}
  \bibinfo{person}{Martin Riedmiller}.} \bibinfo{year}{2013}\natexlab{}.
\newblock \showarticletitle{Playing atari with deep reinforcement learning}.
\newblock \bibinfo{journal}{\emph{arXiv preprint arXiv:1312.5602}}
  (\bibinfo{year}{2013}).
\newblock


\bibitem[Mnih et~al\mbox{.}(2015)]%
        {mnih2015human}
\bibfield{author}{\bibinfo{person}{Volodymyr Mnih}, \bibinfo{person}{Koray
  Kavukcuoglu}, \bibinfo{person}{David Silver}, \bibinfo{person}{Andrei~A
  Rusu}, \bibinfo{person}{Joel Veness}, \bibinfo{person}{Marc~G Bellemare},
  \bibinfo{person}{Alex Graves}, \bibinfo{person}{Martin Riedmiller},
  \bibinfo{person}{Andreas~K Fidjeland}, \bibinfo{person}{Georg Ostrovski},
  {et~al\mbox{.}}} \bibinfo{year}{2015}\natexlab{}.
\newblock \showarticletitle{Human-level control through deep reinforcement
  learning}.
\newblock \bibinfo{journal}{\emph{nature}} \bibinfo{volume}{518},
  \bibinfo{number}{7540} (\bibinfo{year}{2015}), \bibinfo{pages}{529--533}.
\newblock


\bibitem[Nazari et~al\mbox{.}(2018)]%
        {nazari2018reinforcement}
\bibfield{author}{\bibinfo{person}{Mohammadreza Nazari},
  \bibinfo{person}{Afshin Oroojlooy}, \bibinfo{person}{Lawrence Snyder}, {and}
  \bibinfo{person}{Martin Tak{\'a}c}.} \bibinfo{year}{2018}\natexlab{}.
\newblock \showarticletitle{Reinforcement learning for solving the vehicle
  routing problem}.
\newblock \bibinfo{journal}{\emph{Advances in neural information processing
  systems}}  \bibinfo{volume}{31} (\bibinfo{year}{2018}).
\newblock


\bibitem[Paszke et~al\mbox{.}(2019)]%
        {paszke2019pytorch}
\bibfield{author}{\bibinfo{person}{Adam Paszke}, \bibinfo{person}{Sam Gross},
  \bibinfo{person}{Francisco Massa}, \bibinfo{person}{Adam Lerer},
  \bibinfo{person}{James Bradbury}, \bibinfo{person}{Gregory Chanan},
  \bibinfo{person}{Trevor Killeen}, \bibinfo{person}{Zeming Lin},
  \bibinfo{person}{Natalia Gimelshein}, \bibinfo{person}{Luca Antiga},
  {et~al\mbox{.}}} \bibinfo{year}{2019}\natexlab{}.
\newblock \showarticletitle{Pytorch: An imperative style, high-performance deep
  learning library}.
\newblock \bibinfo{journal}{\emph{Advances in neural information processing
  systems}}  \bibinfo{volume}{32} (\bibinfo{year}{2019}).
\newblock


\bibitem[Patel et~al\mbox{.}(2012)]%
        {patel2012knowledge}
\bibfield{author}{\bibinfo{person}{Sheela Patel}, \bibinfo{person}{Carrie
  Baptist}, {and} \bibinfo{person}{Celine d’Cruz}.}
  \bibinfo{year}{2012}\natexlab{}.
\newblock \showarticletitle{Knowledge is power--informal communities assert
  their right to the city through SDI and community-led enumerations}.
\newblock \bibinfo{journal}{\emph{Environment and Urbanization}}
  \bibinfo{volume}{24}, \bibinfo{number}{1} (\bibinfo{year}{2012}),
  \bibinfo{pages}{13--26}.
\newblock


\bibitem[Roy et~al\mbox{.}(2021)]%
        {roy2021prefixrl}
\bibfield{author}{\bibinfo{person}{Rajarshi Roy}, \bibinfo{person}{Jonathan
  Raiman}, \bibinfo{person}{Neel Kant}, \bibinfo{person}{Ilyas Elkin},
  \bibinfo{person}{Robert Kirby}, \bibinfo{person}{Michael Siu},
  \bibinfo{person}{Stuart Oberman}, \bibinfo{person}{Saad Godil}, {and}
  \bibinfo{person}{Bryan Catanzaro}.} \bibinfo{year}{2021}\natexlab{}.
\newblock \showarticletitle{Prefixrl: Optimization of parallel prefix circuits
  using deep reinforcement learning}. In \bibinfo{booktitle}{\emph{2021 58th
  ACM/IEEE Design Automation Conference (DAC)}}. IEEE,
  \bibinfo{pages}{853--858}.
\newblock


\bibitem[Schlichtkrull et~al\mbox{.}(2018)]%
        {schlichtkrull2018modeling}
\bibfield{author}{\bibinfo{person}{Michael Schlichtkrull},
  \bibinfo{person}{Thomas~N Kipf}, \bibinfo{person}{Peter Bloem},
  \bibinfo{person}{Rianne Van Den~Berg}, \bibinfo{person}{Ivan Titov}, {and}
  \bibinfo{person}{Max Welling}.} \bibinfo{year}{2018}\natexlab{}.
\newblock \showarticletitle{Modeling relational data with graph convolutional
  networks}. In \bibinfo{booktitle}{\emph{European semantic web conference}}.
  Springer, \bibinfo{pages}{593--607}.
\newblock


\bibitem[Schulman et~al\mbox{.}(2017)]%
        {schulman2017proximal}
\bibfield{author}{\bibinfo{person}{John Schulman}, \bibinfo{person}{Filip
  Wolski}, \bibinfo{person}{Prafulla Dhariwal}, \bibinfo{person}{Alec Radford},
  {and} \bibinfo{person}{Oleg Klimov}.} \bibinfo{year}{2017}\natexlab{}.
\newblock \showarticletitle{Proximal policy optimization algorithms}.
\newblock \bibinfo{journal}{\emph{arXiv preprint arXiv:1707.06347}}
  (\bibinfo{year}{2017}).
\newblock


\bibitem[Segler et~al\mbox{.}(2018)]%
        {segler2018planning}
\bibfield{author}{\bibinfo{person}{Marwin~HS Segler}, \bibinfo{person}{Mike
  Preuss}, {and} \bibinfo{person}{Mark~P Waller}.}
  \bibinfo{year}{2018}\natexlab{}.
\newblock \showarticletitle{Planning chemical syntheses with deep neural
  networks and symbolic AI}.
\newblock \bibinfo{journal}{\emph{Nature}} \bibinfo{volume}{555},
  \bibinfo{number}{7698} (\bibinfo{year}{2018}), \bibinfo{pages}{604--610}.
\newblock


\bibitem[Silver et~al\mbox{.}(2016)]%
        {silver2016mastering}
\bibfield{author}{\bibinfo{person}{David Silver}, \bibinfo{person}{Aja Huang},
  \bibinfo{person}{Chris~J Maddison}, \bibinfo{person}{Arthur Guez},
  \bibinfo{person}{Laurent Sifre}, \bibinfo{person}{George Van Den~Driessche},
  \bibinfo{person}{Julian Schrittwieser}, \bibinfo{person}{Ioannis Antonoglou},
  \bibinfo{person}{Veda Panneershelvam}, \bibinfo{person}{Marc Lanctot},
  {et~al\mbox{.}}} \bibinfo{year}{2016}\natexlab{}.
\newblock \showarticletitle{Mastering the game of Go with deep neural networks
  and tree search}.
\newblock \bibinfo{journal}{\emph{nature}} \bibinfo{volume}{529},
  \bibinfo{number}{7587} (\bibinfo{year}{2016}), \bibinfo{pages}{484--489}.
\newblock


\bibitem[Silver et~al\mbox{.}(2017)]%
        {silver2017mastering}
\bibfield{author}{\bibinfo{person}{David Silver}, \bibinfo{person}{Julian
  Schrittwieser}, \bibinfo{person}{Karen Simonyan}, \bibinfo{person}{Ioannis
  Antonoglou}, \bibinfo{person}{Aja Huang}, \bibinfo{person}{Arthur Guez},
  \bibinfo{person}{Thomas Hubert}, \bibinfo{person}{Lucas Baker},
  \bibinfo{person}{Matthew Lai}, \bibinfo{person}{Adrian Bolton},
  {et~al\mbox{.}}} \bibinfo{year}{2017}\natexlab{}.
\newblock \showarticletitle{Mastering the game of go without human knowledge}.
\newblock \bibinfo{journal}{\emph{nature}} \bibinfo{volume}{550},
  \bibinfo{number}{7676} (\bibinfo{year}{2017}), \bibinfo{pages}{354--359}.
\newblock


\bibitem[UN-Habitat(2004)]%
        {un2004challenge}
\bibfield{author}{\bibinfo{person}{UN-Habitat}.}
  \bibinfo{year}{2004}\natexlab{}.
\newblock \showarticletitle{The challenge of slums: global report on human
  settlements 2003}.
\newblock \bibinfo{journal}{\emph{Management of Environmental Quality: An
  International Journal}} \bibinfo{volume}{15}, \bibinfo{number}{3}
  (\bibinfo{year}{2004}), \bibinfo{pages}{337--338}.
\newblock


\bibitem[UN-Habitat(2014)]%
        {un2014practical}
\bibfield{author}{\bibinfo{person}{UN-Habitat}.}
  \bibinfo{year}{2014}\natexlab{}.
\newblock \bibinfo{title}{A Practical Guide to Designing, Planning, and
  Executing Citywide Slum Upgrading Programmes}.
\newblock
\newblock


\bibitem[Vaswani et~al\mbox{.}(2017)]%
        {vaswani2017attention}
\bibfield{author}{\bibinfo{person}{Ashish Vaswani}, \bibinfo{person}{Noam
  Shazeer}, \bibinfo{person}{Niki Parmar}, \bibinfo{person}{Jakob Uszkoreit},
  \bibinfo{person}{Llion Jones}, \bibinfo{person}{Aidan~N Gomez},
  \bibinfo{person}{{\L}ukasz Kaiser}, {and} \bibinfo{person}{Illia
  Polosukhin}.} \bibinfo{year}{2017}\natexlab{}.
\newblock \showarticletitle{Attention is all you need}.
\newblock \bibinfo{journal}{\emph{Advances in neural information processing
  systems}}  \bibinfo{volume}{30} (\bibinfo{year}{2017}).
\newblock


\bibitem[Veli{\v{c}}kovi{\'c} et~al\mbox{.}(2018)]%
        {velivckovic2018graph}
\bibfield{author}{\bibinfo{person}{Petar Veli{\v{c}}kovi{\'c}},
  \bibinfo{person}{Guillem Cucurull}, \bibinfo{person}{Arantxa Casanova},
  \bibinfo{person}{Adriana Romero}, \bibinfo{person}{Pietro Li{\`o}}, {and}
  \bibinfo{person}{Yoshua Bengio}.} \bibinfo{year}{2018}\natexlab{}.
\newblock \showarticletitle{Graph Attention Networks}. In
  \bibinfo{booktitle}{\emph{International Conference on Learning
  Representations}}.
\newblock


\bibitem[Wang et~al\mbox{.}(2023)]%
        {wang2023automated}
\bibfield{author}{\bibinfo{person}{Dongjie Wang}, \bibinfo{person}{Yanjie Fu},
  \bibinfo{person}{Kunpeng Liu}, \bibinfo{person}{Fanglan Chen},
  \bibinfo{person}{Pengyang Wang}, {and} \bibinfo{person}{Chang-Tien Lu}.}
  \bibinfo{year}{2023}\natexlab{}.
\newblock \showarticletitle{Automated urban planning for reimagining city
  configuration via adversarial learning: quantification, generation, and
  evaluation}.
\newblock \bibinfo{journal}{\emph{ACM Transactions on Spatial Algorithms and
  Systems}} \bibinfo{volume}{9}, \bibinfo{number}{1} (\bibinfo{year}{2023}),
  \bibinfo{pages}{1--24}.
\newblock


\bibitem[Wang et~al\mbox{.}(2020a)]%
        {wang2020reimagining}
\bibfield{author}{\bibinfo{person}{Dongjie Wang}, \bibinfo{person}{Yanjie Fu},
  \bibinfo{person}{Pengyang Wang}, \bibinfo{person}{Bo Huang}, {and}
  \bibinfo{person}{Chang-Tien Lu}.} \bibinfo{year}{2020}\natexlab{a}.
\newblock \showarticletitle{Reimagining city configuration: Automated urban
  planning via adversarial learning}. In \bibinfo{booktitle}{\emph{Proceedings
  of the 28th international conference on advances in geographic information
  systems}}. \bibinfo{pages}{497--506}.
\newblock


\bibitem[Wang et~al\mbox{.}(2021a)]%
        {wang2021deep}
\bibfield{author}{\bibinfo{person}{Dongjie Wang}, \bibinfo{person}{Kunpeng
  Liu}, \bibinfo{person}{Pauline Johnson}, \bibinfo{person}{Leilei Sun},
  \bibinfo{person}{Bowen Du}, {and} \bibinfo{person}{Yanjie Fu}.}
  \bibinfo{year}{2021}\natexlab{a}.
\newblock \showarticletitle{Deep human-guided conditional variational
  generative modeling for automated urban planning}. In
  \bibinfo{booktitle}{\emph{2021 IEEE international conference on data mining
  (ICDM)}}. IEEE, \bibinfo{pages}{679--688}.
\newblock


\bibitem[Wang et~al\mbox{.}(2021b)]%
        {wang2021reinforced}
\bibfield{author}{\bibinfo{person}{Dongjie Wang}, \bibinfo{person}{Pengyang
  Wang}, \bibinfo{person}{Kunpeng Liu}, \bibinfo{person}{Yuanchun Zhou},
  \bibinfo{person}{Charles~E Hughes}, {and} \bibinfo{person}{Yanjie Fu}.}
  \bibinfo{year}{2021}\natexlab{b}.
\newblock \showarticletitle{Reinforced imitative graph representation learning
  for mobile user profiling: An adversarial training perspective}. In
  \bibinfo{booktitle}{\emph{Proceedings of the AAAI Conference on Artificial
  Intelligence}}, Vol.~\bibinfo{volume}{35}. \bibinfo{pages}{4410--4417}.
\newblock


\bibitem[Wang et~al\mbox{.}(2022b)]%
        {wang2022human}
\bibfield{author}{\bibinfo{person}{Dongjie Wang}, \bibinfo{person}{Lingfei Wu},
  \bibinfo{person}{Denghui Zhang}, \bibinfo{person}{Jingbo Zhou},
  \bibinfo{person}{Leilei Sun}, {and} \bibinfo{person}{Yanjie Fu}.}
  \bibinfo{year}{2022}\natexlab{b}.
\newblock \showarticletitle{Human-instructed Deep Hierarchical Generative
  Learning for Automated Urban Planning}.
\newblock \bibinfo{journal}{\emph{arXiv preprint arXiv:2212.00904}}
  (\bibinfo{year}{2022}).
\newblock


\bibitem[Wang et~al\mbox{.}(2020b)]%
        {wang2020representation}
\bibfield{author}{\bibinfo{person}{Meng-Xiang Wang},
  \bibinfo{person}{Wang-Chien Lee}, \bibinfo{person}{Tao-Yang Fu}, {and}
  \bibinfo{person}{Ge Yu}.} \bibinfo{year}{2020}\natexlab{b}.
\newblock \showarticletitle{On representation learning for road networks}.
\newblock \bibinfo{journal}{\emph{ACM Transactions on Intelligent Systems and
  Technology (TIST)}} \bibinfo{volume}{12}, \bibinfo{number}{1}
  (\bibinfo{year}{2020}), \bibinfo{pages}{1--27}.
\newblock


\bibitem[Wang et~al\mbox{.}(2022a)]%
        {wang2022task}
\bibfield{author}{\bibinfo{person}{Song Wang}, \bibinfo{person}{Kaize Ding},
  \bibinfo{person}{Chuxu Zhang}, \bibinfo{person}{Chen Chen}, {and}
  \bibinfo{person}{Jundong Li}.} \bibinfo{year}{2022}\natexlab{a}.
\newblock \showarticletitle{Task-adaptive few-shot node classification}. In
  \bibinfo{booktitle}{\emph{Proceedings of the 28th ACM SIGKDD Conference on
  Knowledge Discovery and Data Mining}}. \bibinfo{pages}{1910--1919}.
\newblock


\bibitem[Wang et~al\mbox{.}(2018)]%
        {wang2018nervenet}
\bibfield{author}{\bibinfo{person}{Tingwu Wang}, \bibinfo{person}{Renjie Liao},
  \bibinfo{person}{Jimmy Ba}, {and} \bibinfo{person}{Sanja Fidler}.}
  \bibinfo{year}{2018}\natexlab{}.
\newblock \showarticletitle{Nervenet: Learning structured policy with graph
  neural networks}. In \bibinfo{booktitle}{\emph{Proceedings of the
  International Conference on Learning Representations, Vancouver, BC,
  Canada}}, Vol.~\bibinfo{volume}{30}.
\newblock


\bibitem[Weru(2004)]%
        {weru2004community}
\bibfield{author}{\bibinfo{person}{Jane Weru}.}
  \bibinfo{year}{2004}\natexlab{}.
\newblock \showarticletitle{Community federations and city upgrading: the work
  of Pamoja Trust and Muungano in Kenya}.
\newblock \bibinfo{journal}{\emph{Environment and Urbanization}}
  \bibinfo{volume}{16}, \bibinfo{number}{1} (\bibinfo{year}{2004}),
  \bibinfo{pages}{47--62}.
\newblock


\bibitem[Wesolowski and Eagle(2010)]%
        {wesolowski2010parameterizing}
\bibfield{author}{\bibinfo{person}{Amy Wesolowski} {and}
  \bibinfo{person}{Nathan Eagle}.} \bibinfo{year}{2010}\natexlab{}.
\newblock \showarticletitle{Parameterizing the dynamics of slums}. In
  \bibinfo{booktitle}{\emph{2010 AAAI Spring Symposium Series}}.
\newblock


\bibitem[Wu et~al\mbox{.}(2019)]%
        {wu2019simplifying}
\bibfield{author}{\bibinfo{person}{Felix Wu}, \bibinfo{person}{Amauri Souza},
  \bibinfo{person}{Tianyi Zhang}, \bibinfo{person}{Christopher Fifty},
  \bibinfo{person}{Tao Yu}, {and} \bibinfo{person}{Kilian Weinberger}.}
  \bibinfo{year}{2019}\natexlab{}.
\newblock \showarticletitle{Simplifying graph convolutional networks}. In
  \bibinfo{booktitle}{\emph{International conference on machine learning}}.
  PMLR, \bibinfo{pages}{6861--6871}.
\newblock


\bibitem[Wu et~al\mbox{.}(2020)]%
        {wu2020learning}
\bibfield{author}{\bibinfo{person}{Ning Wu}, \bibinfo{person}{Xin~Wayne Zhao},
  \bibinfo{person}{Jingyuan Wang}, {and} \bibinfo{person}{Dayan Pan}.}
  \bibinfo{year}{2020}\natexlab{}.
\newblock \showarticletitle{Learning effective road network representation with
  hierarchical graph neural networks}. In \bibinfo{booktitle}{\emph{Proceedings
  of the 26th ACM SIGKDD international conference on knowledge discovery \&
  data mining}}. \bibinfo{pages}{6--14}.
\newblock


\bibitem[Xue et~al\mbox{.}(2022)]%
        {xue2022quantifying}
\bibfield{author}{\bibinfo{person}{Jiawei Xue}, \bibinfo{person}{Nan Jiang},
  \bibinfo{person}{Senwei Liang}, \bibinfo{person}{Qiyuan Pang},
  \bibinfo{person}{Takahiro Yabe}, \bibinfo{person}{Satish~V Ukkusuri}, {and}
  \bibinfo{person}{Jianzhu Ma}.} \bibinfo{year}{2022}\natexlab{}.
\newblock \showarticletitle{Quantifying the spatial homogeneity of urban road
  networks via graph neural networks}.
\newblock \bibinfo{journal}{\emph{Nature Machine Intelligence}}
  \bibinfo{volume}{4}, \bibinfo{number}{3} (\bibinfo{year}{2022}),
  \bibinfo{pages}{246--257}.
\newblock


\bibitem[Yang et~al\mbox{.}(2022)]%
        {yang2022learning}
\bibfield{author}{\bibinfo{person}{Rui Yang}, \bibinfo{person}{Jie Wang},
  \bibinfo{person}{Zijie Geng}, \bibinfo{person}{Mingxuan Ye},
  \bibinfo{person}{Shuiwang Ji}, \bibinfo{person}{Bin Li}, {and}
  \bibinfo{person}{Feng Wu}.} \bibinfo{year}{2022}\natexlab{}.
\newblock \showarticletitle{Learning Task-relevant Representations for
  Generalization via Characteristic Functions of Reward Sequence
  Distributions}.
\newblock \bibinfo{journal}{\emph{arXiv preprint arXiv:2205.10218}}
  (\bibinfo{year}{2022}).
\newblock


\bibitem[Ying et~al\mbox{.}(2018)]%
        {ying2018graph}
\bibfield{author}{\bibinfo{person}{Rex Ying}, \bibinfo{person}{Ruining He},
  \bibinfo{person}{Kaifeng Chen}, \bibinfo{person}{Pong Eksombatchai},
  \bibinfo{person}{William~L Hamilton}, {and} \bibinfo{person}{Jure Leskovec}.}
  \bibinfo{year}{2018}\natexlab{}.
\newblock \showarticletitle{Graph convolutional neural networks for web-scale
  recommender systems}. In \bibinfo{booktitle}{\emph{Proceedings of the 24th
  ACM SIGKDD international conference on knowledge discovery \& data mining}}.
  \bibinfo{pages}{974--983}.
\newblock


\bibitem[Yun et~al\mbox{.}(2019)]%
        {yun2019graph}
\bibfield{author}{\bibinfo{person}{Seongjun Yun}, \bibinfo{person}{Minbyul
  Jeong}, \bibinfo{person}{Raehyun Kim}, \bibinfo{person}{Jaewoo Kang}, {and}
  \bibinfo{person}{Hyunwoo~J Kim}.} \bibinfo{year}{2019}\natexlab{}.
\newblock \showarticletitle{Graph transformer networks}.
\newblock \bibinfo{journal}{\emph{Advances in neural information processing
  systems}}  \bibinfo{volume}{32} (\bibinfo{year}{2019}).
\newblock


\bibitem[Zong et~al\mbox{.}(2022)]%
        {zong2022rbg}
\bibfield{author}{\bibinfo{person}{Zefang Zong}, \bibinfo{person}{Hansen Wang},
  \bibinfo{person}{Jingwei Wang}, \bibinfo{person}{Meng Zheng}, {and}
  \bibinfo{person}{Yong Li}.} \bibinfo{year}{2022}\natexlab{}.
\newblock \showarticletitle{RBG: Hierarchically Solving Large-Scale Routing
  Problems in Logistic Systems via Reinforcement Learning}. In
  \bibinfo{booktitle}{\emph{Proceedings of the 28th ACM SIGKDD Conference on
  Knowledge Discovery and Data Mining}}. \bibinfo{pages}{4648--4658}.
\newblock


\end{thebibliography}

\newpage
\section*{Appendix}
\appendix

\section{Research Methods}
\subsection{Markov Decision Process}\label{app::mdp}
We propose a DRL model to solve the sequential decision-making problem, where an intelligent agent learns to automatically select locations for road segments by interacting with a slum planning environment, as shown in Figure \ref{fig::rl}.
From the perspective of DRL, the problem can be expressed as a Markov Decision Process (MDP), which contains the following critical components:
\begin{itemize}[leftmargin=*]
    \item \textbf{States} describe the current conditions of the slum, including both static and dynamic features for places and roads.
    \item \textbf{Actions} indicate the selected locations of new road segments.
    \item \textbf{Rewards} provide feedback for road planning actions, which consider the connectivity, travel distance, and construction cost to obtain a comprehensive evaluation.
    \item \textbf{Transitions} express the dynamic changes of the slum, such as the changes of segments from \textit{candidates} to \textit{roads}, and the resulting changes in accessibility and travel distance.
\end{itemize}

\subsection{GNN for Road Planning}\label{app::gnn_comparison}
Existing tasks in which GNN has been proven successful tend to focus more on nodes of the graph, such as learning node representations for node classification~\cite{kipf2016semi,hamilton2017inductive,velivckovic2018graph,wu2019simplifying} and link prediction~\cite{schlichtkrull2018modeling,he2020lightgcn,ying2018graph}.
However, in road planning for slums, not only nodes but also edges and faces need to be taken into consideration, and these topological elements are interdependent with each other.
Specifically, the directly operated decision objects are edges on the graph, which are planned as roads to connect different parts of the slum and have different construction costs.
In addition, the effect of road planning is reflected in the faces of the graph, since they represent places in the slum, of which the accessibility and travel distance are optimized.
Furthermore, both of the above two metrics are calculated on the planned road network, which mainly consists of pathways between nodes.
Consequently, existing node-oriented GNN models are not suitable for road planning, due to their insufficient modeling of the complicated topological structure.
In this paper, we propose topology-aware GNN, which can effectively handle the complex topology of slums.

\subsection{Definitions of Topological Features}\label{app::feature}
We design rich features for topological elements on the graph, including nodes, edges and faces.
These features are used as input of the proposed GNN model to learn representations.
We include 11 categories of features as illustrated in Table \ref{tab::feature}.
We now introduce the specific definitions of these features.

\subsubsection*{\textbf{Node Features}}
Nodes represent junctions in a slum, which are \textit{points} in the original geometric space.
We include the following features,
\begin{itemize}[leftmargin=*]
    \item \textbf{Coordinates}: the Cartesian coordinates $(x, y)$ indicating the location of the junction in the slum.
    \item \textbf{Centrality}: the network centrality metrics of the junction.
    We compute four centrality metrics including degree centrality, betweenness centrality~\cite{freeman1977set}, eigenvector centrality~\cite{bonacich1987power} and closeness centrality~\cite{freeman2002centrality}.
    \item \textbf{On Road}: a boolean feature indicating whether the junction is on a road, either external or planned.
    \item \textbf{Road Ratio}: the ratio of the number of adjacent road edges to the total number of adjacent edges.
    It is 0 when \textit{On Road} is \textit{False}.
    \item \textbf{Avg N2N Dis}: the average distance from the node to all other nodes over the constructed road network.
    It is set as a very large value if the node is not on a road.
\end{itemize}

\subsubsection*{\textbf{Edge Features}}
Edges represent paths in a slum, which are \textit{line segments} in the original geometric space.
We include the following features,
\begin{itemize}[leftmargin=*]
    \item \textbf{Cost}: the construction cost of building the segment as a road, which is set as the length of the path.
    \item \textbf{Road}: a boolean feature indicating whether the path is a road or not.
    A road can be an existing external one or a newly planned one.
    \item \textbf{Straightness}: the ratio of the road network distance to the euclidean distance between the two endpoints.
\end{itemize}

\subsubsection*{\textbf{Face Features}}
Faces represent places in a slum, which are \textit{polygons} in the original geometric space.
We include the following features,
\begin{itemize}[leftmargin=*]
    \item \textbf{Connected}: a boolean feature indicating whether the place is connected to the road system.
    \item \textbf{Avg F2F Dis}: the average distance from the place to all other places over the constructed road network.
    It is set as a very large value if the face is not connected to the road system.
    \item \textbf{F2E Dis}: the distance from the place to the external boundaries $E$ of the slum.
\end{itemize}

\subsection{Model Training and Inference}\label{app::model}

\subsubsection{\textbf{Model Training}}
We collect hundreds of rollout trajectories in the replay buffer, and utilize PPO~\cite{schulman2017proximal} to train our policy and value networks.
Specifically, the loss function is composed of policy loss, entropy loss, and value loss.

Policy loss clips the objective to control the change of the policy at each iteration, which encourages the policy to conduct safe exploration.
The loss function is calculated as follows,
\begin{align}
    &r_t(\theta) = \frac{\pi_\theta(a_t | s_t)}{\pi_{\theta_{\text{old}}}(a_t | s_t)}, \\
    &L_p = \text{min}(r_t(\theta)\hat{A}_t, \text{clip}(r_t(\theta), 1 - \epsilon, 1 + \epsilon)\hat{A}_t),
\end{align}
where $\theta$ is the parameters of the policy network, $r_t(\theta)$ is the ratio of action probability of the new policy to the old policy, $\hat{A}_t$ is the advantage value calculated according to rewards at each step, and $\text{clip}$ controls the update not to be too large.

Entropy loss is included to balance between exploitation and exploration, which is calculated as follows,
\begin{equation}
    L_e = \text{Entropy}[\text{Prob}(a_1), \cdots, \text{Prob}(a_{\mathcal{E}})],
\end{equation}
where $\mathcal{E}$ is the total number of edges, and $\text{Prob}$ is obtained by the policy network according to (\ref{eq::prob}).

We adopt mean squared error (MSE) between the predicted return and the groundtruth return as the value loss to supervise the value network, which is calculated as follow,
\begin{equation}
    L_v = \text{MSE}(\hat{r}_t, R_t),
\end{equation}
where $R_t$ is the real return value, and $\hat{r}_t$ is estimated by the value network according to (\ref{eq::value}).

We integrate the above loss functions through weighted sum,
\begin{equation}
    L = L_p + \beta L_e + \gamma L_v,
\end{equation}
where $\beta$ and $\gamma$ are hyper-parameters in our model.

To ease the modeling training, we fix $W_{f \rightarrow e}^{(l+1)}$ in (\ref{eq::face2edge}) and $W_{e \rightarrow e}^{(l+1)}$ in (\ref{eq::edge2edge}) as identity matrix $I$.
Making the two linear transformation layers learnable can further increase the capability of our model, and we leave it for future work.

\subsubsection{\textbf{Model Inference}}
During model training, actions are sampled according to the selection probability in (\ref{eq::prob}).
Once a well-trained model is obtained, we take actions greedily according to the probability, \textit{i.e.}, the most likely action is taken to layout road segments at each step.
Specifically, we use the policy network to compute the probability distribution over different actions and takes the action with the largest probability value,
\begin{equation}
    a = \text{argmax}\{\text{Prob}(a_1), \cdots, \text{Prob}(a_{\mathcal{E}})\}.
\end{equation}
We perform the above fast model inference to generate road plans.

\section{Details of Slum Data}\label{app::data}
\begin{figure}[t]
    \centering
    \vspace{-15px}
    \includegraphics[width=0.99\linewidth]{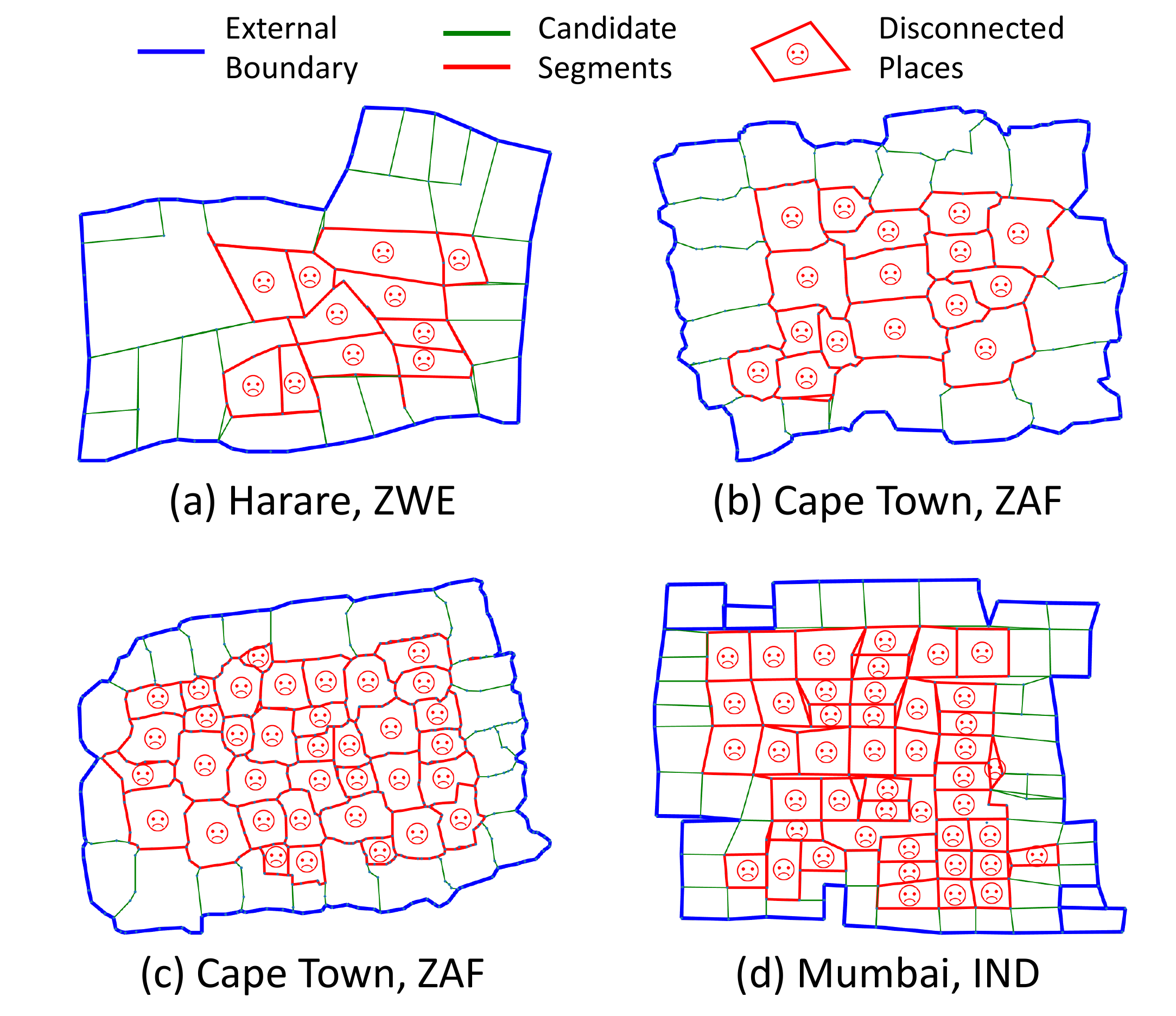}
    \caption{
    Geometrical descriptions of the adopted four slums in (a) Harare, ZWE (b) Cape Town, ZAF (c) Cape Town, ZAF (d) Mumbai, IND.
    All slums suffer from poor accessibility, with a large fraction of places disconnected to the road system.
    Best viewed in color.
    }
    \vspace{-15px}
    \label{fig::app_slum_data}
\end{figure}

As shown in Table \ref{tab::data}, we conduct experiments on four real-world slums from three different countries, including Zimbabwe (ZWE), South Africa (ZAF) and India (IND).
The specific locations of the four slums are Epworth (Harare, ZWE), Khayelitsha (Cape Town, ZAF) and Phule Nagar (Mumbai, IND).
The digital maps and the geometrical descriptions of places and roads for the four slums are publicly released by~\cite{brelsford2018toward}.
Figure \ref{fig::app_slum_data} illustrate the geometrical descriptions of the four slums.

\section{Implementation Details}\label{app::implementation}

\subsection{Baselines}\label{app::baseline}
We implement Random, Greedy and GA-G methods and integrate it into our framework.
Each method outputs the score for each edge $e_{ij}$ that is transformed to selection probability with (\ref{eq::prob}).

\subsubsection*{\textbf{Random}}
The score for each edge is obtained in a fully random way,
\begin{equation}
    s_{ij} = \text{Random}().
\end{equation}

\subsubsection*{\textbf{Greedy-A}}
The score of the edge which connects the most disconnected places is set as 1, while others are set as 0,
\begin{equation}
    s_{ij} = \mathbbm{1}[e_{ij} = \text{argmax}\{\sum_{f \in F_{n_j}}{(1 - \text{Connected}(f))}\}].
\end{equation}

\subsubsection*{\textbf{Greedy-C}}
The score of the edge with the least construction cost is set as 1, while others are set as 0,
\begin{equation}
    s_{ij} = \mathbbm{1}[\mathcal{C}_{e_{ij}}=\text{min}\{\mathcal{C}\}].
\end{equation}

\subsubsection*{\textbf{MST}}
We utilize Kruskal's algorithm~\cite{kruskal1956shortest} to grow a minimum spanning tree on a transformed graph, where nodes represent places in the slum, and edges represent candidate road segments with their weights as construction cost.

\subsubsection*{\textbf{GA-G}}
We utilize a vector and compute inner product between this vector and the edge attributes to obtain the score for each edge,
\begin{equation}
    s_{ij} = <v_{GA-G}, A_{e_{ij}}>,
\end{equation}
where $v_{GA-G}$ is set as the genes.
We initialize a random population and perform crossover and mutation to evolve for better road plans.

For the \textit{\textbf{GA-S}} baseline, we set the gene as the selection of edges,
\begin{equation}
    g_{GA-S} = [\text{Selected}_1,\cdots,\text{Selected}_{\mathcal{E}}],
\end{equation}
where the sum of $g_{GA-S}$ is the road planning budget $K$.
We perform crossover which is swapping between different solutions, as well as mutation to evolve the population.
The fitness function of a solution is defined as a weighted sum of accessibility, travel distance and construction costs.
For the \textit{\textbf{HS-MC}}~\cite{brelsford2018toward} baseline, we use the original codes\footnote{\url{https://github.com/brelsford/topology}} released by the authors.

Generative models are not applicable to the problem of road planning for slums, due to the significant differences in planning constraints, planning targets, planning budget, and data representation.
Still, we exhausted all efforts to adapt two typical generative models \cite{fang2022incorporating,kempinska2019modelling} based on GAN~\cite{goodfellow2020generative} and VAE~\cite{kingma2013auto} to accomplish the task of road planning for slums. 
Specifically, we have to conduct the following tedious adaptions which make the road planning process far from automation.
\begin{itemize}[leftmargin=*]
    \item First, we have to manually move the generated roads to their nearest candidate segments to avoid destroying existing houses in the slum.
    \item Second, we have to manually remove some of the generated roads to avoid breaking the planning budget.
    \item Third, we have to manually transform the generated raster representation to a vector representation to obtain a faithful solution.
\end{itemize}

\subsection{Hyper-parameters of Our Model}\label{app::our_model}
We implement the proposed model with PyTorch~\cite{paszke2019pytorch}, and release the codes at at \textcolor{blue}{\url{https://github.com/tsinghua-fib-lab/road-planning-for-slums}}.
We tune the hyper-parameters of our model carefully and list the adopted values in Table \ref{tab::app_hyper}.

\begin{table}[t]
\caption{Designed features for topological elements.}
\label{tab::app_hyper}
\vspace{-10px}
\begin{tabular}{ccc}
\toprule
\textbf{Category}     & \textbf{Hyper-parameter} & \textbf{Value} \\
\midrule
\multirow{4}{*}{\bf{Network}} & GNN layer & 2   \\
& GNN node dimension & 16 \\
& Policy Head $\verb|MLP|_{p}$ & [32, 1] \\
& Value Head $\verb|MLP|_{v}$ & [32, 32, 1]  \\
\hline
\multirow{4}{*}{\bf{PPO}} & gamma & 0.995 \\
& tau & 0 \\
& Entropy Loss $\beta$ & 0.01  \\ 
& Value Loss $\gamma$ & 0.5  \\ 
\hline
\multirow{3}{*}{\bf{Train}} & optimizer & Adam \\
& weight decay & 0 \\
& learning rate & 0.0004 \\
\bottomrule
\end{tabular}
\vspace{-10px}
\end{table}

\section{Differences with Road Planning for Common Regions}\label{app::difference_generative}
In the paper, we explained why current city-level approaches cannot handle the micro-level road planning within a slum. 
In this section, we would like to further clarify the differences between road planning for common regions and slums, as well as why generative models are not applicable for this problem.

\textbf{First}, the \textbf{planning constraints} make road planning for slums a problem quite different from road planning for common regions. 
Specifically, road planning for common regions is more like \textit{painting on a white paper}, where roads can grow at almost anywhere since there are no existing land functionalities in the planned region. 
As a consequence, a street network can be generated in a process similar to image generation, and generative models such as GAN~\cite{goodfellow2020generative} can be adopted. 
However, \textit{slum is not a white paper, where existing houses determine the possible forms of the road network}. 
As stated in Section \ref{sec::prob}, planned roads are not allowed to pass through the middle of places to minimize disruption to the slums, thus the candidate locations are restricted to the spacing between places. 
In other words, \textit{road planning for slums is more a decision-making task than a generation task}, where the road plans are actually obtained by choosing a fraction of segments from a candidate set of segments, rather than painting on a white paper. Therefore, it is quite different from road planning for common regions, and generative models such as GAN~\cite{goodfellow2020generative} and VAE~\cite{kingma2013auto} are not applicable in road planning for slums.

\textbf{Second}, the \textbf{planning targets} of road planning for slums are very different from road planning for common regions. 
In road planning for slums, as stated in Section \ref{sec::prob}, we aim to \textit{make every place in the slum connected to the road system (universal connectivity), reduce travel distance between different places, and maintain a affordable construction cost}. 
On the contrary, in road planning for common regions, especially through generative methods such as \cite{fang2022incorporating,kempinska2019modelling}, the target is to \textit{learn the patterns of street network based on surrounding context}. 
Meanwhile, for a given slum, we need to add roads in it instead of recovering a masked region as in \cite{fang2022incorporating}. 
For the model design, as the planning process is to select a fraction of segments from a candidate set to improve accessibility rather than learning certain road network patterns, generative models are not appropriate for the task of road planning for slums.

\textbf{Third}, the \textbf{planning budget} also makes road planning for slums a different problem. 
Specifically, we need to upgrade the slum under a given budget, which can be the number of road segments to be added or the monetary cost of the planned road. 
Therefore, \textit{it is more a sequential decision-making problem, where an agent adds a segment or a path in each step, and terminates the process when it runs out of the budget}. 
Generative models such as \cite{fang2022incorporating}, generate road plans in one pass, which makes it very difficult to satisfy the requirements of planning budgets.

\textbf{Fourth}, with respect to \textbf{data representation}, to obtain a faithful solution to road planning for slums, a slum and roads in it need to be described with an \textit{accurate vector representation (points, linestrings and polygons with exact coordinates), rather than a vague raster representation (image pixels)}. 
Therefore, for the model design, a model that outputs discrete actions selecting segments is more appropriate than a model that generates an image such as \cite{fang2022incorporating}.

In summary, road planning for slums is a very different problem from road planning for common regions, in terms of planning constraints, planning targets, planning budget, and data representation. 
These differences all directly influence the model design, making generative models not applicable to this problem. 

\section{Additional Results}
\subsection{Generated Road Plans}\label{app::plan}
\begin{figure*}[h]
    \centering
    \includegraphics[width=0.9\linewidth]{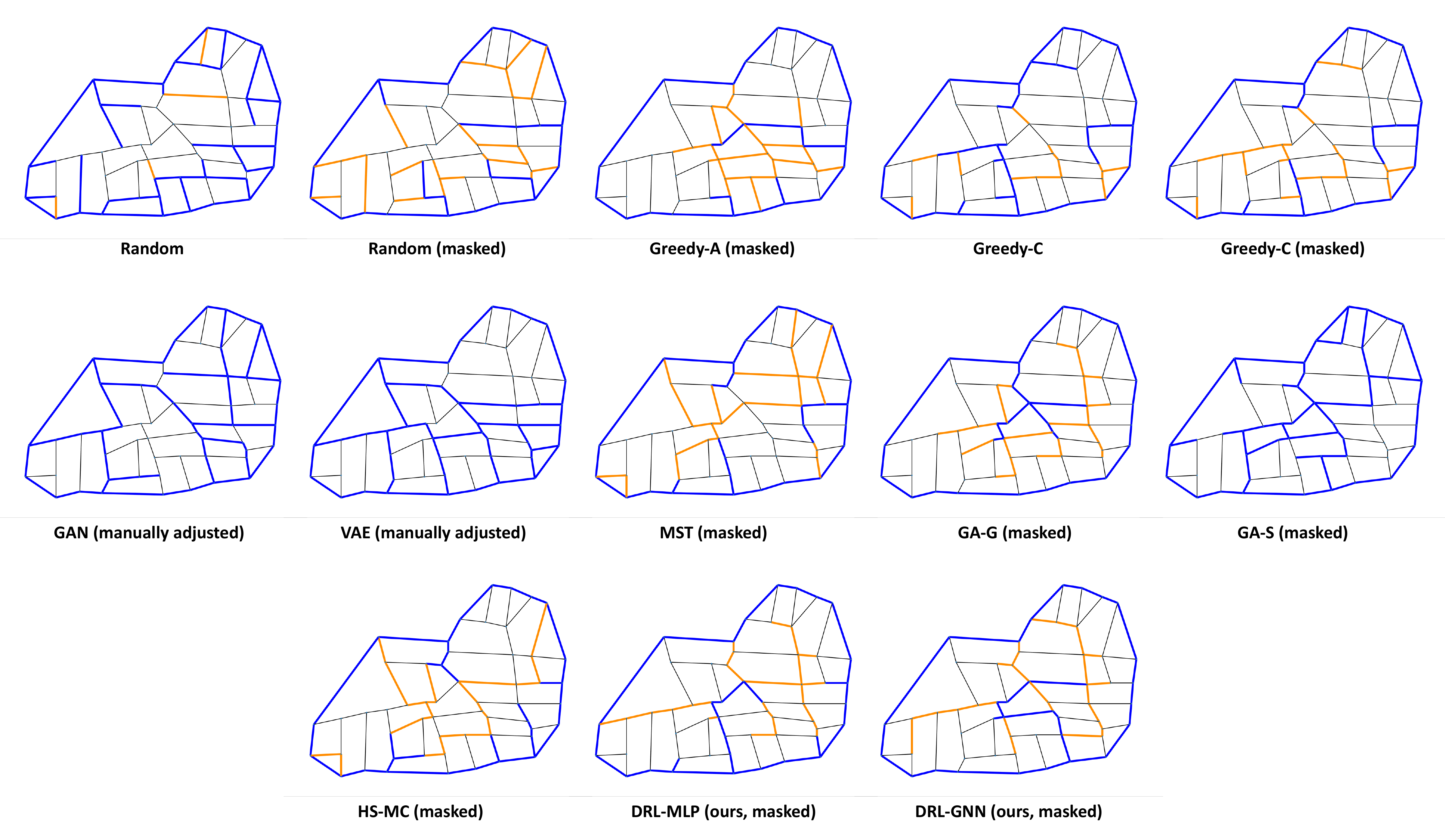}
    \vspace{-5px}
    \caption{
    The generated road plans of all methods for the slum in Harare, ZWE.
    Red polygons are remaining disconnected places after road planning.
    Blue segments except for external boundaries are planned roads in stage \uppercase\expandafter{\romannumeral1}.
    Orange segments are planned roads in stage \uppercase\expandafter{\romannumeral2}.
    Best viewed in color.
    }
    \vspace{-5px}
    \label{fig::app_all_plan_harare}
\end{figure*}

\begin{figure*}[h]
    \centering
    \includegraphics[width=0.9\linewidth]{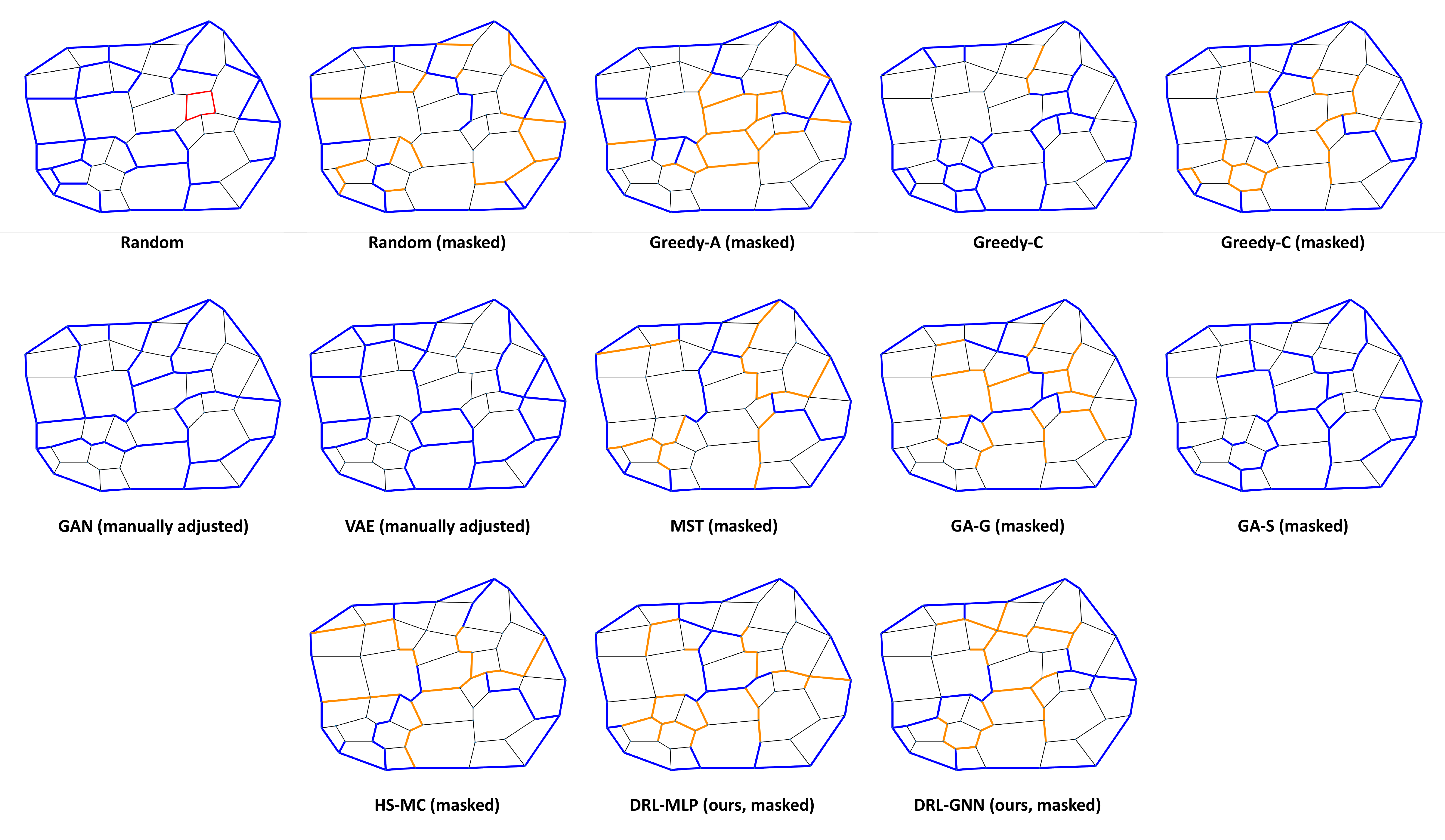}
    \vspace{-5px}
    \caption{
    The generated road plans of all methods for the slum in Cape Town, ZAF.
    Red polygons are remaining disconnected places after road planning.
    Blue segments except for external boundaries are planned roads in stage \uppercase\expandafter{\romannumeral1}.
    Orange segments are planned roads in stage \uppercase\expandafter{\romannumeral2}.
    Best viewed in color.
    }
    \vspace{-5px}
    \label{fig::app_all_plan_capetown1}
\end{figure*}

\begin{figure*}[h]
    \centering
    \includegraphics[width=0.9\linewidth]{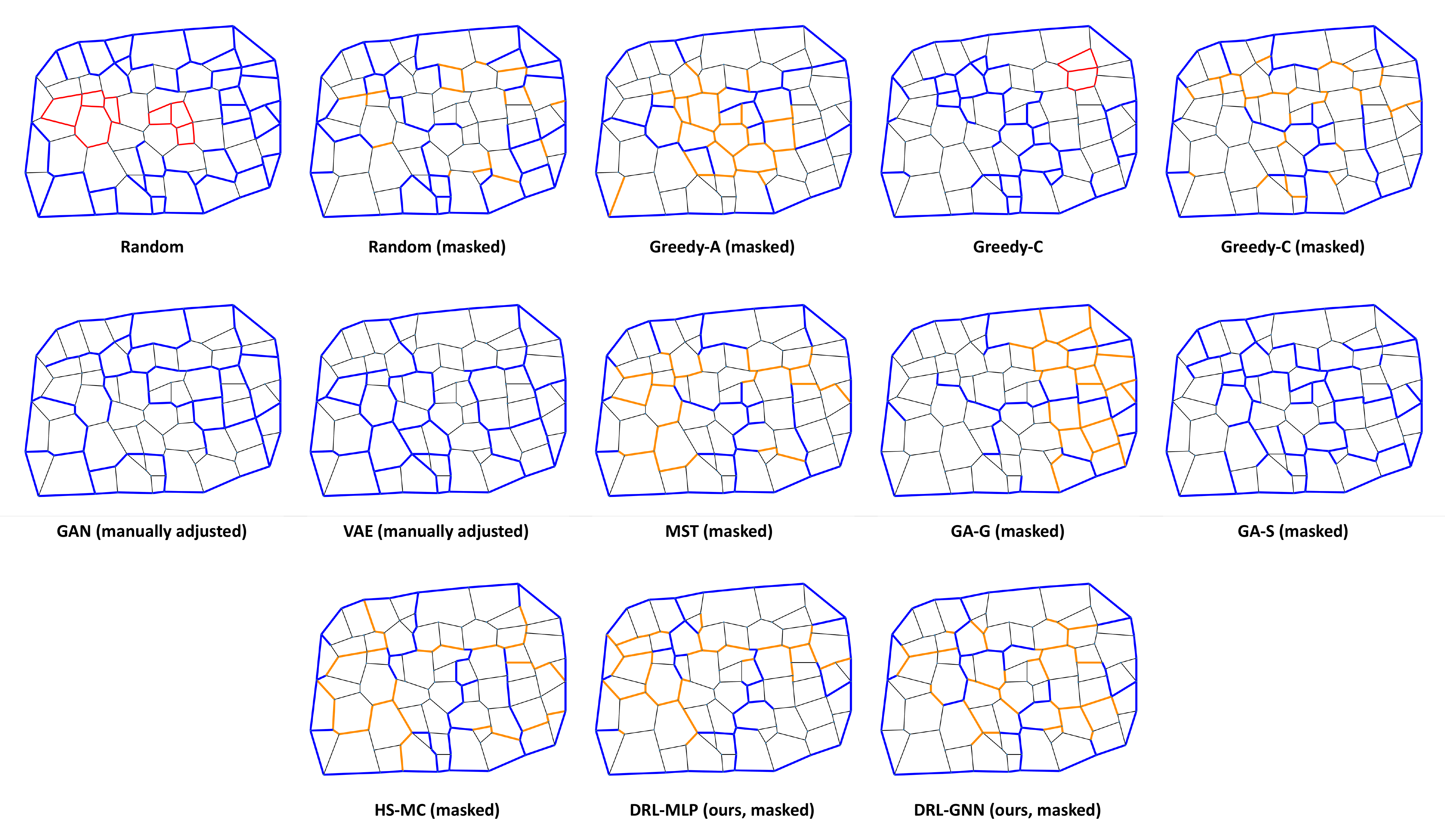}
    \vspace{-5px}
    \caption{
    The generated road plans of all methods for the slum in Cape Town, ZAF.
    Red polygons are remaining disconnected places after road planning.
    Blue segments except for external boundaries are planned roads in stage \uppercase\expandafter{\romannumeral1}.
    Orange segments are planned roads in stage \uppercase\expandafter{\romannumeral2}.
    Best viewed in color.
    }
    \vspace{-5px}
    \label{fig::app_all_plan_capetown2}
\end{figure*}

\begin{figure*}[h]
    \centering
    \includegraphics[width=0.9\linewidth]{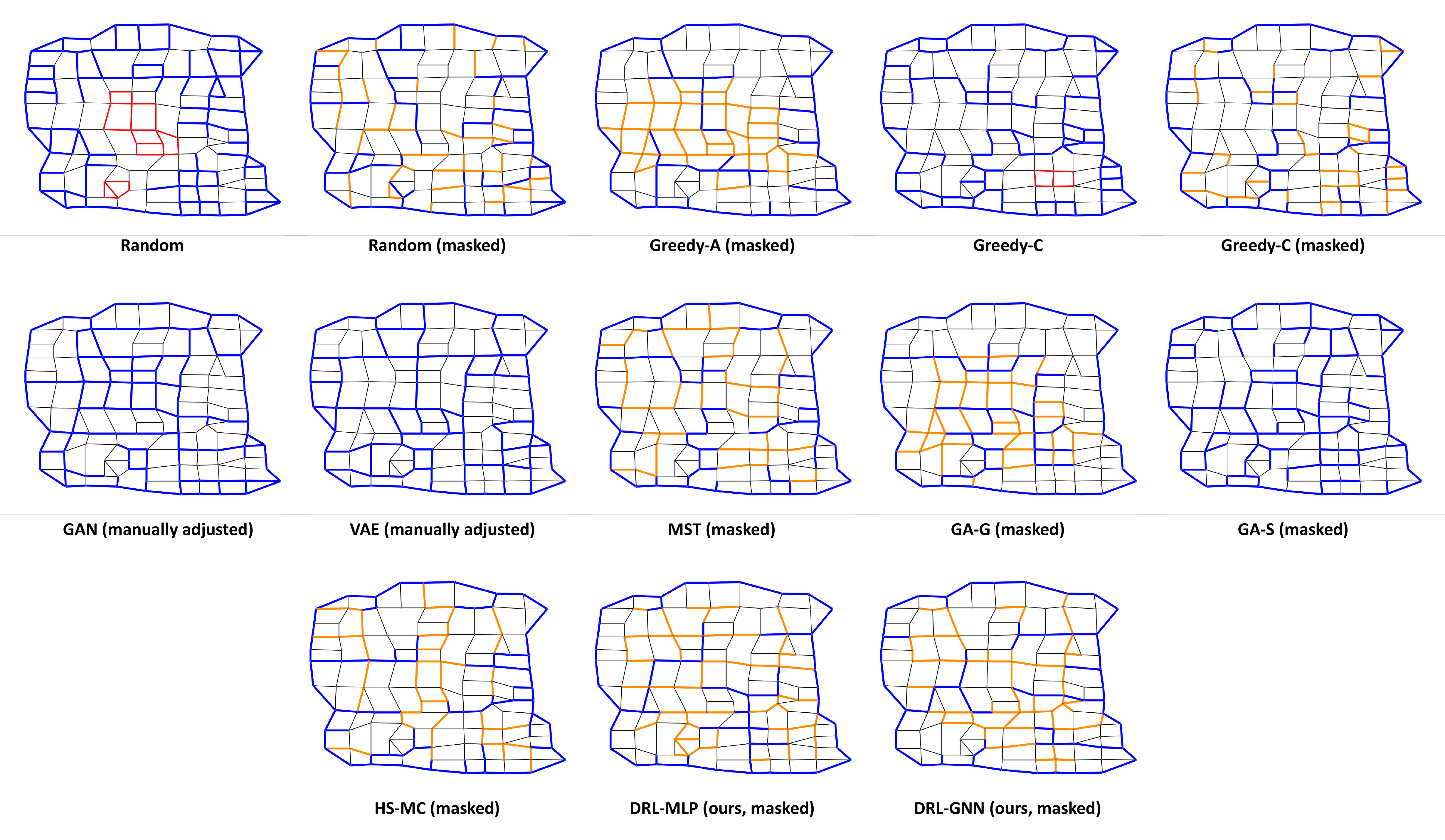}
    \vspace{-5px}
    \caption{
    The generated road plans of all methods for the slum in Mumbai, IND.
    Red polygons are remaining disconnected places after road planning.
    Blue segments except for external boundaries are planned roads in stage \uppercase\expandafter{\romannumeral1}.
    Orange segments are planned roads in stage \uppercase\expandafter{\romannumeral2}.
    Best viewed in color.
    }
    \vspace{-5px}
    \label{fig::app_all_plan_mumbai}
\end{figure*}

In Section \ref{sec::perf_comp}, we show the road planning performance of different methods for the four slum, as well as the generated road plans of GA-G, HS-MC and our DRL-GNN for the slum in Cape Town, ZAF.
Our proposed model significantly outperforms other baselines for all slums, improving their accessibility and travel distance with lower construction costs.
We provide the complete generated road plans of 13 methods on 4 slums in Figure \ref{fig::app_all_plan_harare}-\ref{fig::app_all_plan_mumbai}.

\subsection{Convergence of Our Models}\label{app::convergence}
\begin{figure}[t]
    \centering
    \includegraphics[width=0.99\linewidth]{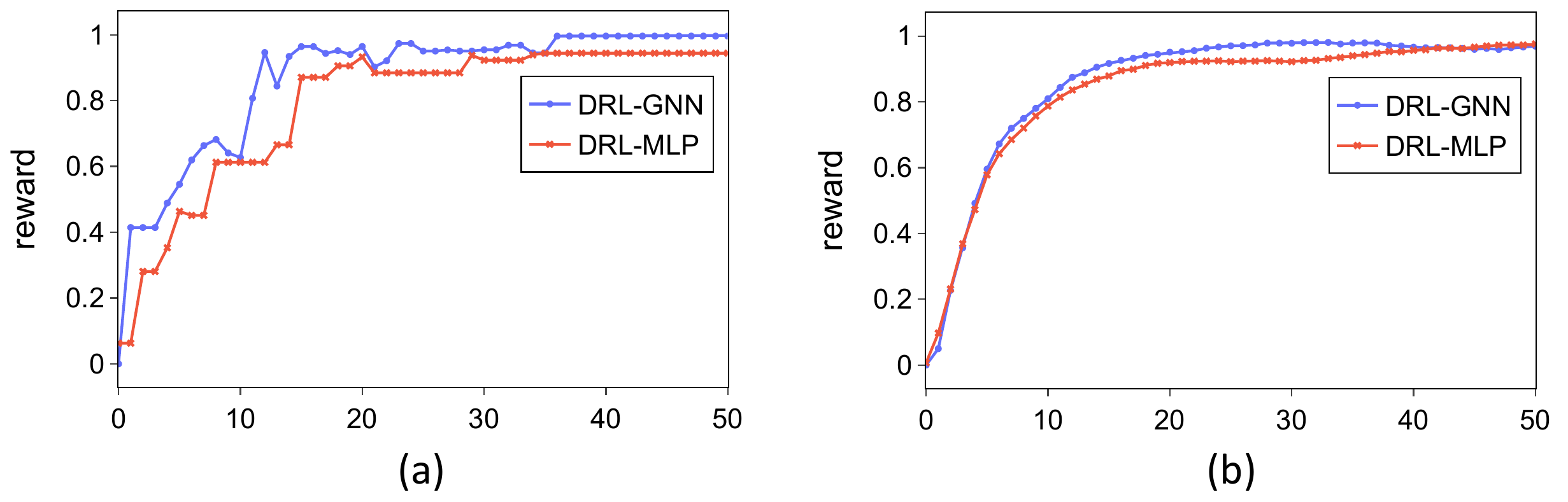}
    \vspace{-10px}
    \caption{
    Normalized episodic reward of DRL-MLP and DRL-GNN after each iteration of training for the slums in (a) Harare, ZWE (b) Cape Town, ZAF.
    Best viewed in color.
    }
    \label{fig::app_convergence}
    \vspace{-10px}
\end{figure}

With the specially designed multi-objective policy optimization, our proposed models can efficiently learn the skills of road plans within only about 2 hours of training on a single server with a single Nvidia GeForce 2080Ti GPU.
We show the normalized episodic reward of our models after each iteration of training for the slums in Harare, ZWE and Cape Town, ZAF in Figure \ref{fig::app_convergence}.
We can observe that both DRL-MLP and DRL-GNN can reach convergence within less than 100 iterations, which is highly efficient.

\subsection{Using Cost as Planning Budget}\label{app::cost_budget}

\begin{table*}[t]
\caption{Road planning performance comparison using construction cost as planning budget.
Lower is better.
}
\vspace{-10px}
\label{tab::cost_budget}
\begin{tabular}{c|c|cc|cc|cc|cc}
\toprule
\multirow{2}{*}{\textbf{Budget (SC)}} & \multirow{2}{*}{\textbf{Method}} & \multicolumn{2}{c|}{\textbf{Harare, ZWE}} & \multicolumn{2}{c|}{\textbf{Cape Town, ZAF (A)}} & \multicolumn{2}{c|}{\textbf{Cape Town, ZAF (B)}} & \multicolumn{2}{c}{\textbf{Mumbai, IND}} \\
                                & & \textbf{UC} & \textbf{AD} & \textbf{UC}   & \textbf{AD}   & \textbf{UC}   & \textbf{AD}  & \textbf{UC}   & \textbf{AD} \\
\midrule
\multirow{3}{*}{Low} & HS-MC & NO & INF & NO & INF & NO & INF & NO & INF    \\
& DRL-GNN (ours) & \bf{YES} & \bf{1.26} & \bf{YES} & \bf{2.47} & \bf{YES} & \bf{2.77} & \bf{YES} & \bf{3.34} \\
& impr\% v.s. HS-MC & -100\% & -100\% & -100\% & -100\% & -100\% & -100\% & -100\% & -100\%  \\
\hline
\multirow{3}{*}{Medium} & HS-MC & YES & 0.70 & YES & 1.41 & YES & 1.96 & YES & 2.09    \\
& DRL-GNN (ours) & YES & \bf{0.65} & YES & \bf{1.24} & YES & \bf{1.76} & YES & \bf{1.85} \\
& impr\% v.s. HS-MC & - & -7.1\% & - & -12.1\% & - & -10.2\% & - & -11.5\%  \\
\hline
\multirow{3}{*}{High} & HS-MC & YES & 0.64 & YES & 1.21 & YES & 1.68 & YES & 1.71    \\
& DRL-GNN (ours) & YES & \bf{0.52} & YES & \bf{1.02} & YES & \bf{1.44} & YES & \bf{1.55} \\
& impr\% v.s. HS-MC & - & -18.8\% & - & -15.7\% & - & -14.3\% & - & -9.4\%  \\
\bottomrule
\end{tabular}
\vspace{-5px}
\end{table*}

\begin{table*}[t]
\caption{Road planning performance comparison using construction cost as planning budget.
Lower is better.
}
\vspace{-10px}
\label{tab::cost_budget_train}
\begin{tabular}{c|cc|cc|cc|cc}
\toprule
\multirow{2}{*}{\textbf{Method}} & \multicolumn{2}{c|}{\textbf{Harare, ZWE}} & \multicolumn{2}{c|}{\textbf{Cape Town, ZAF (A)}} & \multicolumn{2}{c|}{\textbf{Cape Town, ZAF (B)}} & \multicolumn{2}{c}{\textbf{Mumbai, IND}} \\
& \textbf{UC} & \textbf{AD} & \textbf{UC}   & \textbf{AD}   & \textbf{UC}   & \textbf{AD}  & \textbf{UC}   & \textbf{AD} \\
\midrule
HS-MC & YES & 0.64 & YES & 1.21 & YES & 1.68 & YES & 1.71    \\
DRL-GNN (ours) & YES & \bf{0.51} & YES & \bf{0.94} & YES & \bf{1.44} & YES & \bf{1.53} \\
impr\% v.s. HS-MC & - & -20.3\% & - & -22.3\% & - & -14.3\% & - & -10.5\%  \\
\bottomrule
\end{tabular}
\vspace{-5px}
\end{table*}

In the paper, we use the number of road segments to be planned, $K$, as the planning budget.
In fact, there are other acceptable and feasible budgets for road planning, such as the construction cost.
Fortunately, our model is not limited to a designated type of budget and is still applicable if the budget is changed to the construction cost. 
We will explain it below in detail with experimental results.

In fact, our model can generate decent road plans with optimal accessibility and minimal cost under a budget specified in terms of the number of roads, or the construction cost, or any other types of budget. 
From the perspective of reinforcement learning where the intelligent road planning agent interacts with a simulated slum environment, the budget is only used by the environment but not the agent. 
Specifically, the environment uses the budget value to determine whether to continue or terminate the planning process. 
Under both types of budget, the agent always builds one segment at each step, until the environment claims that it has run out of budget. 
Therefore, when changing the budget to the construction cost, we only need to redefine the termination logic of the environment, and we can simply keep the model as it is. 
To show this point, we conducted supplementary experiments by utilizing the trained model under the original budget of the number of road segments, and directly testing it with the different budget of construction cost. 
We even changed the budget to different values of construction cost without any modification to the model. 
The results are shown in Table \ref{tab::cost_budget}. 
As the construction cost becomes budget now, the SC metric will no longer make sense, thus we only report AD and whether universal connectivity (UC) is achieved.

We can observe that the results under the budget of construction cost is consistent with the results in the paper which is trained under the budget of the number of road segments. 
With low budget of construction cost, HS-MC can not achieve universal connectivity while our DRL-GNN can successfully connect all places in the slum. 
For all slums under all different budget of construction cost, the travel distance of our approach is significantly lower than that of HS-MC.

Besides the above results of direct evaluation, we also train a new model with the same high budget of construction cost in Table \ref{tab::cost_budget} (we only redefine the termination logic of the environment, and use the same model structure). 
The results are shown in Table \ref{tab::cost_budget_train}. 
Similarly, we only report UC and AD, because SC does not make sense.
We can observe that the results are consistent with previous findings, where our proposed DRL-GNN method significantly outperforms HS-MC with over 10\% relative improvements.

In summary, our proposed model can deal with different definitions of the planning budget without any modification. 
Specifically, we can simply keep the model unchanged, and only redefine the termination logic of the environment.
The results under different planning budgets are quite stable, showing the consistent advantages of our proposed method.

\subsection{More Results on Graph Modeling}\label{app::graph_model}
We have shown that DRL-GNN achieves better final road planning performance compared with a DRL model without the generic graph modeling, DRL-MLP, in Table \ref{tab::overall}.
Despite the superior evaluation performance, the graph modeling makes it easier to discover places with poor accessibility, which in turn facilitates more efficient search in the action space.
As a consequence, DRL-GNN is much more efficient considering training samples than DRL-MLP.
Table \ref{tab::app_graph} illustrates the number of iterations utilized to train DRL-GNN and DRL-MLP until convergence for the four slums.
We can observe that DRL-GNN requires much fewer iterations to converge for all slums.
Notably, for the largest slum in Mumbai, IND, our DRL-GNN model uses over 70\% fewer iterations to achieve better road planning performance than DRL-MLP.
The superior sample efficiency of our DRL-GNN model indicate that it can generate decent road plans in a short time, which is critical for real-world slum upgrading with a large solution space.

\begin{table}[t]
\caption{The number of iterations for model convergence.}
\label{tab::app_graph}
\begin{tabular}{c|cc|c}
\toprule
\bf{Method}         & \bf{DRL-MLP} & \bf{DRL-GNN} & \bf{impr\%}  \\
\midrule
Harare, ZWE    & 91      & 57      & -37.4\% \\
Cape Town, ZAF & 79      & 31      & -60.8\% \\
Cape Town, ZAF & 87      & 40      & -54.0\% \\
Mumbai, IND    & 82      & 23      & -72.0\% \\
\bottomrule
\end{tabular}
\end{table}

\subsection{More Results on Topological Features}\label{app::input_features}
\begin{figure}[t]
    \centering
    \includegraphics[width=0.99\linewidth]{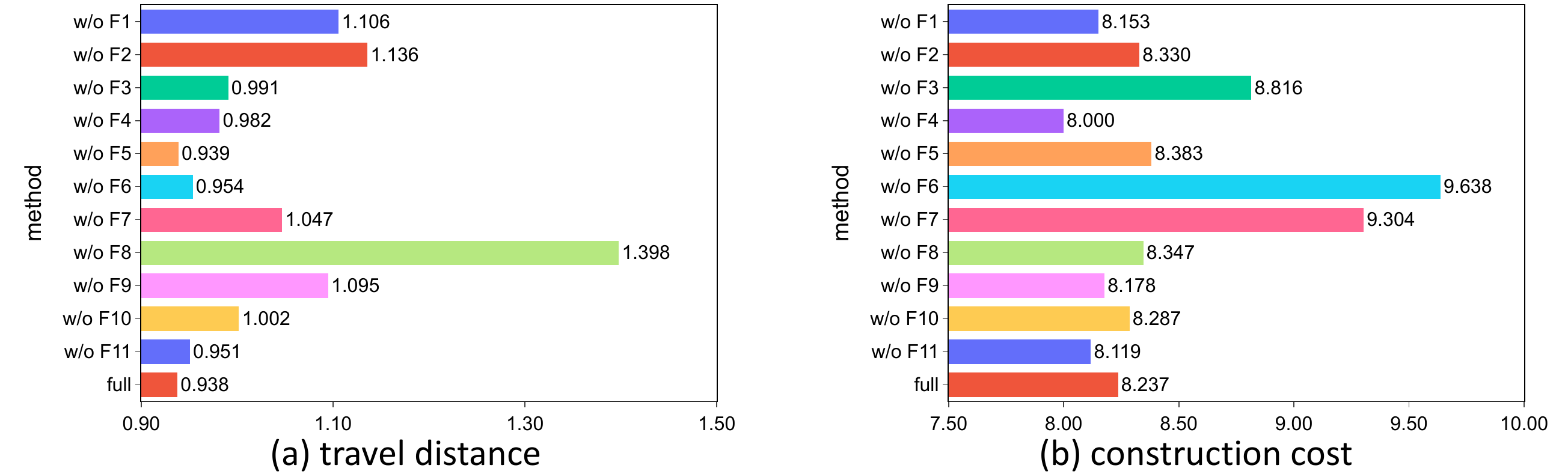}
    \vspace{-10px}
    \caption{
    Road planning performance of our model and its variants by removing different features with respect to (a) travel distance (b) construction cost for the slum in Cape Town, ZAF.
    Features are Coordinates (F1), Centrality (F2), On Road (F3), Road Ratio (F4), Avg N2N Dis (F5), Cost (F6), Road (F7), Straightness (F8), Connected (F9), Avg F2F Dis (F10) and F2E Dis (F11).
    Best viewed in color.
    }
    \label{fig::app_all_features}
    \vspace{-10px}
\end{figure}

We investigate the effect of different topological features by setting the values as 0 and evaluating their corresponding performance.
In Section \ref{sec::ablation}, we investigate three typical features, namely \textit{Centrality}, \textit{Road} and \textit{Straightness}.
We demonstrate the results of removing other features in Figure \ref{fig::app_all_features}.
We can observe that all features contribute to the effect of road planning with respect to travel distance, and removing any of them will bring about a deterioration.
Among these features, Straightness (F8), Centrality (F2), Coordinates (F1), and Connected (F9) are the most important features, and removing them lead to an over 16.7\% increase in travel distance.
These four features, especially Straightness, directly reflect the current conditions of travel distance, and can be utilized by our model to detect long detours in the slum.
In terms of construction cost, the Cost feature (F6) is the most important which is consistent with intuition.
All these features contain rich topological information about the accessibility, travel distance and constructions costs of roads and places in the slum.
We adopt the 11 categories of features since they reflect topological information and they are all easy to compute.
Furthermore, we believe more features can be designed and included into our model, which may lead to better road planning performance.

\subsection{More Results on Action Mask}\label{app::action_mask}

\begin{figure}[t]
    \centering
    \includegraphics[width=0.99\linewidth]{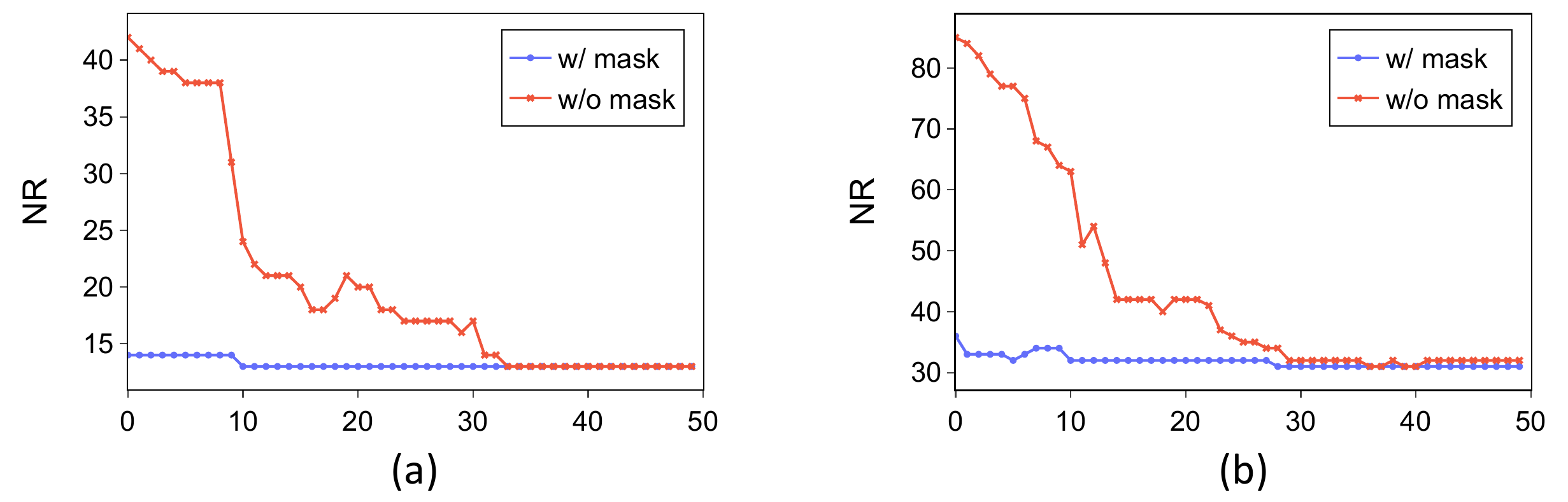}
    \vspace{-10px}
    \caption{
    Accessibility after each iteration of training of our model with and without the action mask for the two slums (a and b) in Cape Town, ZAF.
    Best viewed in color.
    }
    \label{fig::app_mask_iteration}
    \vspace{-10px}
\end{figure}

We have shown that the masked policy optimization method can improve the final road planning metrics in Section \ref{sec::ablation}.
Besides better evaluation performance, the action mask can block out actions of low quality, such as the roads segments which connect places that are already connected.
In other words, the action mask guides our model to explore in a \textit{sub action space} which is smaller and contains much fewer undesirable actions, instead of the original enormous one.
Therefore, our model can learn skills of road planning faster with the help of the action mask.
We illustrate the convergence curve of accessibility of our DRL-GNN model with and without the action mask in Figure \ref{fig::app_mask_iteration}, which is the number of road segments (NR) consumed to achieve universal connectivity after each iteration of training.
We can observe that although both methods eventually achieve the same lowest NR, using the action mask can make our DRL-GNN model converge much faster than optimizing it without the mask.
Specifically, without the action mask, the DRL-GNN model tends to try those actions of low quality in the early stages of training, thus it requires much larger number of road segments to achieve universal connectivity.
In the first 20 iterations, the NR is over 20 and 40 for the two slums respectively.
On the contrary, adding the action mask help the DRL-GNN model quickly discover road segments that can significantly improve the accessibility.
Thus the NR is getting very close to optimal after only 10 iterations of training, which is 3 times faster than not using the action mask.

\begin{table*}[t]
\caption{Road planning performance comparison of our proposed framework with different network structures.
Lower is better.
}
\vspace{-10px}
\label{tab::gt}
\begin{tabular}{c|ccc|ccc|ccc|ccc}
\toprule
\multirow{2}{*}{\textbf{Method}} & \multicolumn{3}{c|}{\textbf{Harare, ZWE}} & \multicolumn{3}{c|}{\textbf{Cape Town, ZAF (A)}} & \multicolumn{3}{c|}{\textbf{Cape Town, ZAF (B)}} & \multicolumn{3}{c}{\textbf{Mumbai, IND}} \\
                                 & \textbf{NR} & \textbf{AD} & \textbf{SC} & \textbf{NR}   & \textbf{AD}   & \textbf{SC}  & \textbf{NR}   & \textbf{AD}   & \textbf{SC} & \textbf{NR}   & \textbf{AD}   & \textbf{SC}  \\
\midrule
DRL-MLP (ours, masked) & 11 & 0.52 & \bf{4.38} & \uline{14} & \uline{0.96} & \uline{8.28} & \uline{32} & 1.57 & \uline{15.66} & \uline{31} & \uline{1.52} & 22.93    \\
DRL-GNN (ours, masked) & \uline{9} & \bf{0.50} & 4.60 & \bf{13} & \bf{0.93} & \bf{8.24} & \bf{31} & \uline{1.51} & \bf{15.62} & \bf{29} & \bf{1.51} & \uline{22.82} \\
DRL-GT (ours, masked) & \bf{8} & \uline{0.51} & \uline{4.47} & \bf{13} & \bf{0.93} & 8.64 & \uline{32} & \bf{1.38} & 16.97 & \bf{29} & 1.55 & \bf{21.89} \\
\bottomrule
\end{tabular}
\vspace{-5px}
\end{table*}

\subsection{Hyper-parameter Study}\label{app::hyper}

We investigate three key hyper-parameters of our model in this section, which are the number of GNN layers, the dimenstion of GNN representations and the reward weights $(\alpha_1, \alpha_2)$.

\begin{figure}[t]
    \centering
    \includegraphics[width=0.99\linewidth]{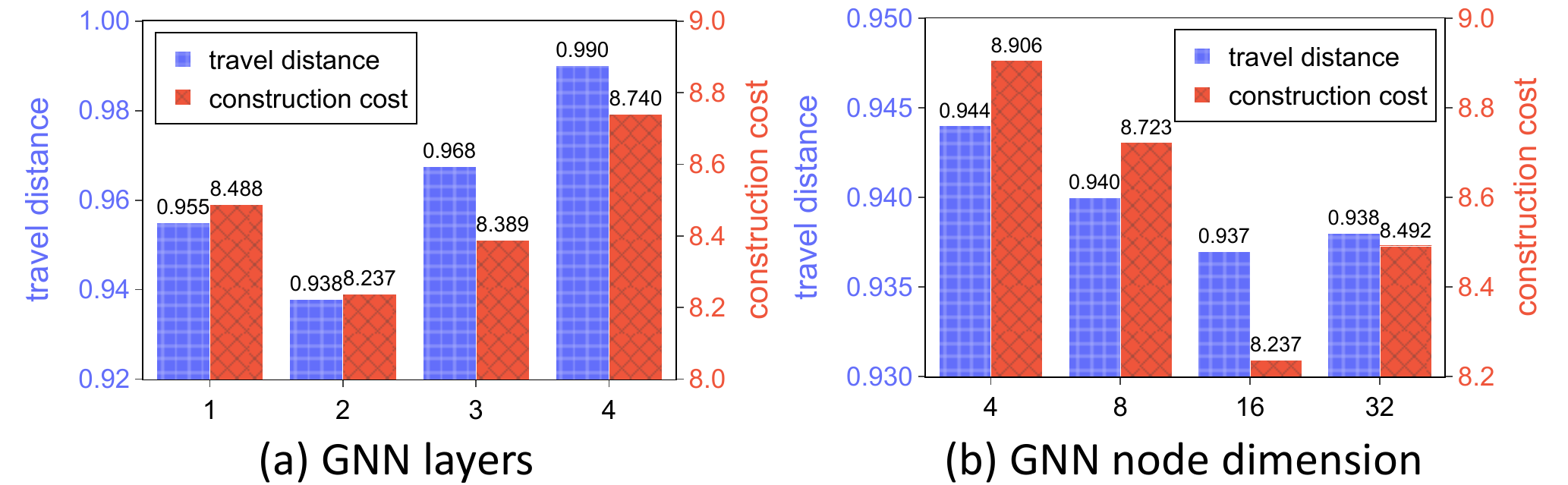}
    \vspace{-10px}
    \caption{
    Performance of DRL-GNN under different values of (a) GNN layers (b) GNN node dimension for the slum in Cape Town, ZAF.
    Best viewed in color.
    }
    \label{fig::app_hyper_gnn}
    \vspace{-10px}
\end{figure}

\noindent\textbf{GNN Layers.}
We design a topology-aware message passing mechanism with node-to-edge propagation, face-to-edge propagation, edge self-propagation and edge embedding broadcast in one GNN layer.
Multiple GNN layers can be stacked to expand the perception fields of each node and edge on the graph, making the learned representations absorb information of multi-hop neighbors.
However, too many GNN layers may lead to over-smoothing and deteriorate the final performance~\cite{chen2020measuring}.
We train our model with different number of GNN layers, and Figure \ref{fig::app_hyper_gnn}(a) demonstrates the results.
Consistent with our expectation, using 2 GNN layers achieves the best performance, while using too few (1) or too many (4) GNN layers both fail to achieve effective road planning.
Particularly, using 4 GNN layers increase travel distance and construction costs by over 5.5\% and 6.1\%, respectively, which is even worse than DRL-MLP.

\noindent\textbf{GNN Embedding Dimension.}
The embedding dimension decides the expressive power of the learned representations.
With too low embedding dimension, the representations for roads and places cannot well capture the topological information.
On the contrary, with too high embedding dimension, the model may suffer from overfitting.
We investigate the performance of of model with different embedding dimensions, as shown in Figure \ref{fig::app_hyper_gnn}(b).
Setting embedding dimension as 16 outperforms other variants of our model.
Increasing it to 32 will worsen construction costs by about 3.1\%, and decreasing it to 4 will make travel distance and construction costs worse by about 0.7\% and 8.1\%.
We thus set embedding dimension as 16 in our experiments.

\begin{figure}[t]
    \centering
    \includegraphics[width=0.99\linewidth]{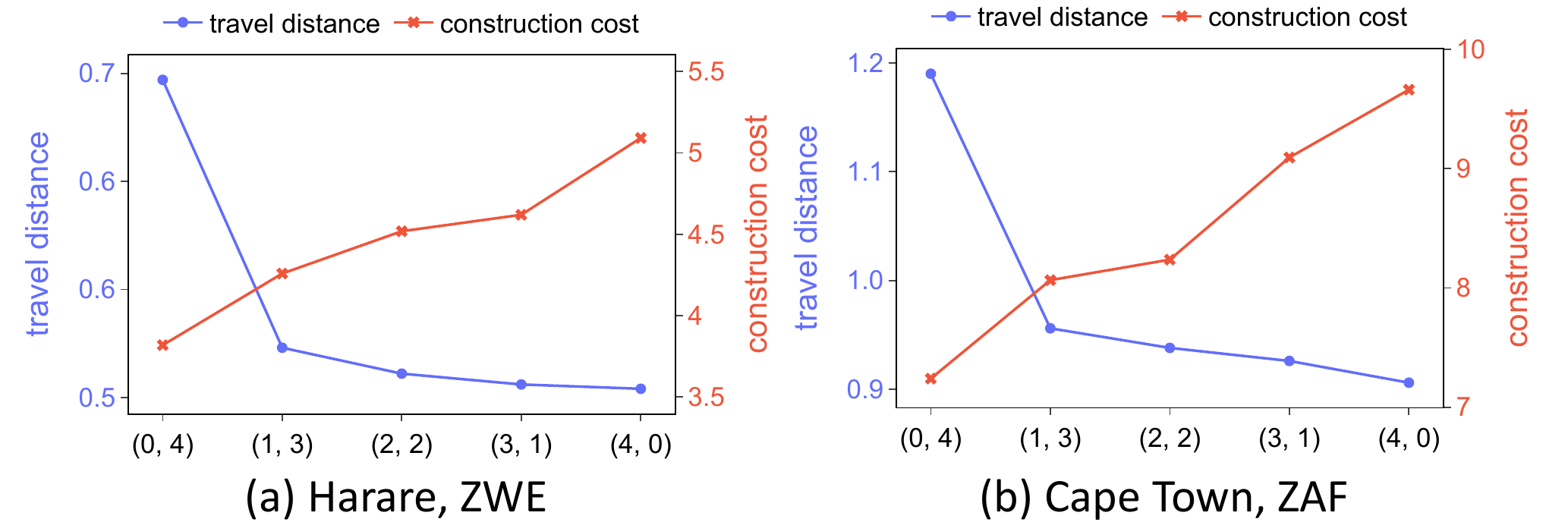}
    \vspace{-10px}
    \caption{
    Performance of DRL-GNN under different values of reward weights $(\alpha_1, \alpha_2)$ for the slum in (a) Harare, ZWE (b) Cape Town, ZAF.
    Best viewed in color.
    }
    \label{fig::app_hyper_reward}
\end{figure}

\noindent\textbf{Reward Weight.}
In (\ref{eq::reward_2}), we add hyper-parameters, $\alpha_1$ and $\alpha_2$, which determine the weight for different rewards.
By altering the ratio of $\alpha_1$ to $\alpha_2$, we can easily optimize our model to emphasize travel distance or construction costs.
Figure \ref{fig::app_hyper_reward} shows the planning performance of our model under different ratios of $\alpha_1$ to $\alpha_2$.
We can observe that using a larger ratio $\alpha_1:\alpha_2$ can effectively reduce the travel distance, while requiring higher construction costs. 
Similarly, a smaller $\alpha_1:\alpha_2$ can prioritize construction costs over travel distance.
Through changing the value of $\alpha_1:\alpha_2$, our model can achieve smooth trade-off between different metrics, which is much beneficial for practical slum upgrading since it usually involves comparison between different road plans.

\subsection{Results of Graph Transformer}\label{app::graph_transformer}

In the paper, we chose to use topology-aware GNN for building our DRL-GNN model, which has been shown to be effective for modeling the topological relationships among different elements in the slum. 
In fact, other graph-based neural network structures can be much useful and deserve thorough exploration. 
Fortunately, our proposed method is flexible and can be easily integrated with more advanced network structures.
In particular, we implement graph transformer (GT)~\cite{yun2019graph}, which replaces the average pooling in (\ref{eq::embed_avg}) with a self-attention module~\cite{vaswani2017attention}. 
The results are shown in Table \ref{tab::gt}.
We can observe that the DRL-GT model achieved competitive performance, indicating the promising potential of our proposed framework to include more advanced neural network structures. 
For instance, our framework can be further extended by incorporating recent progress made in GNN and RL.

\end{document}